\documentclass{elsarticle}
\usepackage{enumerate}
\usepackage{graphicx}
\usepackage{natbib}
\usepackage{marvosym}
\usepackage{amssymb}
\usepackage{booktabs} 
\usepackage{multirow}
\usepackage{hyperref}
\usepackage{color}
\usepackage{amsmath}
\usepackage[linesnumbered,ruled]{algorithm2e}
\SetKwProg{Fn}{Function}{}{end}\SetKwFunction{FRecurs}{FnRecursive}%

\usepackage{lineno,hyperref}
\modulolinenumbers[5]

\begin{document}

\begin{frontmatter}

\title{MOEA/D with Angle-based Constrained Dominance Principle for Constrained Multi-objective Optimization Problems}


\author[mymainaddress]{Zhun Fan}
\author[mymainaddress]{Yi Fang}

\author[mymainaddress]{Wenji Li}

\author[mysecondaryaddress]{Xinye Cai\corref{mycorrespondingauthor}}
\cortext[mycorrespondingauthor]{Corresponding author}
\ead{xinye@nuaa.edu.cn}
\author[mythirdaddress]{Caimin Wei}
\author[myfourthaddress]{Erik Goodman}

\address[mymainaddress]{Department of Electronic Engineering, Shantou University, Guangdong, 515063, China}
\address[mysecondaryaddress]{College of Computer Science and Technology, Nanjing University of Aeronautics and Astronautics, Jiangsu, 210016, China}
\address[mythirdaddress]{Department of Mathematics, Shantou University, Guangdong, 515063, China}
\address[myfourthaddress]{BEACON Center for the Study of Evolution in Action, Michigan State University. East Lansing, Michigan, USA}

\begin{abstract}
This paper proposes a novel constraint-handling mechanism named angle-based constrained dominance principle (ACDP) embedded in a decomposition-based multi-objective evolutionary algorithm (MOEA/D) to solve constrained multi-objective optimization problems (CMOPs). To maintain the diversity of the working population, ACDP utilizes the information of the angle of solutions to adjust the dominance relation of solutions during the evolutionary process. This paper uses 14 benchmark instances to evaluate the performance of the MOEA/D with ACDP (MOEA/D-ACDP). Additionally, an engineering optimization problem (which is I-beam optimization problem) is optimized. The proposed MOEA/D-ACDP, and four other decomposition-based CMOEAs, including C-MOEA/D, MOEA/D-CDP, MOEA/D-Epsilon and MOEA/D-SR are tested by the above benchmarks and the engineering application. The experimental results manifest that MOEA/D-ACDP is significantly better than the other four CMOEAs on these test instances and the real-world case, which indicates that ACDP is more effective for solving CMOPs.
\end{abstract}


\end{frontmatter}


\section{Introduction}
\label{intro}
Multi-objective optimization problems (MOPs) involve the optimization of more than one objective function. In the real world, many optimization problems invariably involve a number of constraints and a multiple conflicting objectives. In general, a CMOP can mathematically be described as follows:
\begin{equation}
\label{equ:cmop_definition}
\begin{cases}
\mbox{minimize} &\mathbf{F}(\mathbf{x}) = {(f_{1}(\mathbf{x}),\ldots,f_{m}(\mathbf{x}))} ^ {T} \\
\mbox{subject to} & g_i(\mathbf{x}) \ge 0, i = 1,\ldots,q \\
& h_j(\mathbf{x}) = 0, j= 1,\ldots,p \\
&\mathbf{x} \in{\mathbb{R}^n}
\end{cases}
\end{equation}
where $F(\mathbf{x}) = ({f_1}(\mathbf{x}),{f_2}(\mathbf{x}), \ldots ,{f_m}(\mathbf{x})) ^ T \in \mathbb{R} ^m$ is an $m$-dimensional objective vector, ${g_i}(\mathbf{x}) \ge 0$ is the $i^{th}$ inequality constraint, and ${h_j}(\mathbf{x})=0$ is the $j^{th}$ equality constraint. $\mathbf{x} \in \mathbb{R}^n$ is an $n$-dimensional decision vector. The feasible region $S$ is defined as the set $\{\mathbf{x} | g_i(\mathbf{x}) \ge 0, i = 1,\ldots,q$  and $h_j(\mathbf{x}) = 0, j= 1,\ldots,p\}$.

In CMOPs, there are usually more than one constraint.  To demonstrate the degree of constraint violation, these constraints are commonly summarized into a scalar value as follows:
\begin{eqnarray}
\label{equ:constraint}
\phi(\mathbf{x}) = \sum_{i=1}^{q} |\min(g_i(\mathbf{x}),0)| + \sum_{j = 1}^{p} |h_j(\mathbf{x})|
\end{eqnarray}
when $\phi(\mathbf{x})=0$, the solution $\mathbf{x}$ is feasible, otherwise it is infeasible.

For any two feasible solutions $\mathbf{x}^a\in S$ and $\mathbf{x}^b\in S$ of a CMOP, it can be said that $\mathbf{x}^a$ dominates $\mathbf{x}^b$ if the following condition is met:
\begin{equation}
\label{equ:dominate}
\forall i ~ f_i(\mathbf{x}^a)\le f_i(\mathbf{x}^b)~~ \text{and} ~~   \exists j~  f_j(\mathbf{x}^a)< f_j(\mathbf{x}^b)
\end{equation}
where $i,\ j \in \{1,2,...,m\}$. If there exists a solution $x^*\in S$, which dominates any other solutions in $S$, $x^*$ can be said as a Pareto optimal solution. The set of all Pareto optimal solution belonging to $S$ is called as Pareto set (PS). The set of the mapping vectors of PS in the objective space is named as Pareto front (PF), which can be defined in the form of $PF=\{F(x)|\ x\in PS\}$.

CMOPs are consist of a few objectives and constraints. To solve CMOPs, the constraint-handling technique should be applied to the framework of multi-objective evolutionary algorithm (MOEA).

According to different selection strategies, MOEAs mainly can be classified into three types: (1) Pareto-domination-based; (2) decomposition-based; (3) indicator-based. The typical examples of domination-based MOEAs include NSGA-II \cite{996017}, MOGA \cite{Murata1995MOGA}, PAES-II \cite{corne2001pesa}, SPEA-II \cite{zitzler2001spea2} and NPGA \cite{Horn2002A}. In recent years, decomposition-based MOEAs attract a lot of attention. The most representative algorithm of this type is MOEA/D \cite{Zhang:2007va}. Some variants of MOEA/D include MOEA/D-DE \cite{Li:2009vo}, MOEA/D-M2M \cite{Liu:2014jb}, EAG-MOEA/D \cite{Cai:2015gi} and MOEA/D-SAS \cite{Cai:2016ii}. For indicator-based MOEAs, they use a scalar metric to assist the selection, typical examples of this type are IBEA \cite{Zitzler:2004tm}, SMS-EMOA \cite{Beume2007SMS}, HypE \cite{Bader2011Hype} and FV-MOEA \cite{Jiang:2015hj}.

To solve CMOPs, constraint-handling mechanisms are important. In recent years, many different constraint-handling mechanisms have been proposed \cite{Cai:2013iz, Hu:2013kc}. According to \cite{CoelloCoello20021245},  constraint-handling   techniques   can   be  generally classified into  (1)  penalty  functions;  (2)  special  representations  and operators; (3) repair algorithms; (4) separate objectives and constraints; and (5) hybrid methods.

As a representative objective-constraint separating method, constrained-domination principle (CDP) proposed in \cite{996017} is widely used. It solves CMOPs by treating constraints as the top priority. Stochastic ranking \cite{runarsson2000stochastic}, $\varepsilon$-constrained method \cite{Takahama2006cons} and non-greedy constraint-handling technique \cite{singh2010c} are also in this category.

Currently, CTPs \cite{ deb2001multi} and CFs \cite{zhang2008multiobjective} are the most widely used CMOP test instances. The common characteristic of these benchmarks is that they all have large feasible regions in the objective space. However, constraint-handling mechanisms do not work when the working population falls in feasible regions, so these two series of instances are not actually suitable for testing the effectiveness of constraint-handling mechanisms.

The rest of this paper is organized as follows: Section \ref{sec:2} briefly introduces MOEA/D and four other decomposition-based CMOEAs. Section \ref{sec:3} introduces the details of the angle-based constrained dominance principle embedded in MOEA/D. Section \ref{sec:4} gives comprehensive experimental results of the proposed algorithm MOEA/D-ACDP and four other CMOEAs on LIR-CMOPs and the I-beam optimization problem. Finally, conclusions are made in section \ref{sec:5}.
\section{Relative Work}
\label{sec:2}
\subsection{MOEA/D}
\label{sec:2.1}
In the original framework of MOEA/D \cite{Zhang:2007va}, given a series of uniform distributed weight vectors, a MOP is decomposed into $N$ scalar subproblems (SOPs), and each SOP relates to one solution. In MOEA/D, a set of $N$ uniformly spread weight vectors $\lambda^1,\ldots,\lambda^N$ is initially generated for $N$ subproblems. A weight vector $\lambda^i$ satisfies the following conditions:
\begin{eqnarray}
\sum_{k = 1} ^ {m}\lambda_{k}^{i} = 1\quad \text{and} \quad \lambda_{k}^i \ge 0 \quad  \text{for each}\ k \in \{1,\ldots,m\}.
\label{equ:weight}
\end{eqnarray}

There are several approaches to decompose a MOP into a number of scalar optimization subproblems \cite{Zhang:2007va, miettinen1999nonlinear}. Three decomposition approaches, including weighted sum \cite{miettinen1999nonlinear}, Tchebycheff \cite{miettinen1999nonlinear} and boundary intersection approaches \cite{Zhang:2007va} are commonly used. In this paper, Tchebycheff decomposition method is used in the MOEA/D framework. The $j$-th subproblem is defined as follows:
\begin{eqnarray}
&\nonumber \mbox{minimize} &g^{te}(\mathbf{x}|\lambda,z^{*}) = \max_{1 \le i \le m} \left\{\frac{1}{\lambda_i^j} |f_i(\mathbf{x}) - z_{i}^{*}| \right\}\\
&\mbox{subject to} & \mathbf{x} \in{S}
\label{equ:tchmethod-a}
\end{eqnarray}
where $z^* = (z_1^*,\ldots,z_m^*)$ is the ideal point, and $z_i^* = \min \{f_i(\mathbf{x}) | \mathbf{x} \in S\}$.

\subsection{Decomposition-based CMOEAs}
\label{sec:2.2}
Decomposition-based CMOEAs combine the MOEA/D with different constraint-handling mechanisms. In this paper, we introduce four representative decomposition-based CMOEAs including C-MOEA/D \cite{Asafuddoula2012An}, MOEA/D-CDP \cite{jan2013study}, MOEA/D-Epsilon \cite{Yang2014Epsilon}, and MOEA/D-SR \cite{jan2013study}.
\begin{itemize}
  \item C-MOEA/D \cite{Asafuddoula2012An} uses a variant of the epsilon constraint-handling technique. In this technique, the epsilon level is set to handle constraints according to the constraint violation and the proportion of feasible solutions in the current population. When comparing any two solutions, if overall constraint violations of the solutions are both less than the epsilon level, the one with a better aggregation value dominates the other. Otherwise, the one with a smaller overall constraint violation dominates the other.

  \item MOEA/D-CDP \cite{jan2013study} uses CDP \cite{996017} to judge the domination relationship between two arbitrary solutions. CDP can be simply summarized as following three rules: \\
1) When two feasible solutions are compared, the one with a better aggregation value dominates the other. \\
2) When a feasible solution is compared with an infeasible solution, the feasible solution dominates the infeasible solution. \\
3) When two infeasible solutions are compared, the one with a smaller degree of constraint violation dominates the other.\\
The second and the third rules can be combined as a single rule: When at least one of two compared solutions is infeasible, the one with a smaller degree of constraint violation dominates the other.
  \item MOEA/D-Epsilon \cite{Yang2014Epsilon} uses the original epsilon constraint-handling technique. The epsilon level setting can be referred to \cite{Takahama2006cons}. With the generation counter $K$ increasing, the epsilon level will dynamically decrease.

  \item MOEA/D-SR \cite{jan2013study} embeds the stochastic ranking method (SR) \cite{runarsson2000stochastic} in MOEA/D to deal with constraints. A threshold parameter $r_f \in [0,1]$ is set to balance the selection between the objectives and the constraints in MOEA/D-SR. When comparing two solutions, if a random number in $[0,1]$ is less than $r_f$, the one with a better aggregation value is retained into the next generation. If the random number in $[0,1]$  is greater than $r_f$, the whole framework of MOEA/D-SR is similar to that of MOEA/D-CDP. In the case of $r_f = 0$, MOEA/D-SR is equivalent to MOEA/D-CDP.
\end{itemize}
\section{MOEA/D with Angle-based Constrained Dominance Principle}
\label{sec:3}
In this section, the definition of the proposed ACDP and the effectiveness of this mechanism in MOEA/D are detailed.

\subsection{Angle-based Constrained Dominance Principle}
\label{sec:3.1}
In the CDP approach \cite{996017}, with its three basic rules, the overall constraint violation is the most important factor during the evolutionary process, and some useful information in the infeasible regions tends to be ignored.

The angle between any two solutions in the objective space is useful information, which is related to the similarity between these two solutions. The definition of angle between any two solutions $\mathbf{x}^1$ and $\mathbf{x}^2$ is given as follows:
\begin{eqnarray}
angle(\mathbf{x}^1,\mathbf{x}^2,z^*)\!=\!\arccos\left(\!\frac{(\mathbf F(\mathbf{x}^1)-z^*)^T \cdot (\mathbf F(\mathbf{x}^2)-z^*)}{||\mathbf F(\mathbf{x}^1)-z^*||\cdot||\mathbf F(\mathbf{x}^2)-z^*||}\!\right)
\label{equ:angle-count}
\end{eqnarray}
where $z^* = (z_1^*,\ldots,z_m^*)$ is the ideal point, and $z_i^* = \min \{f_i(\mathbf{x} | \mathbf{x} \in S\}$. $||\cdot||$ is the two-norm of a vector.

\begin{figure*}[htbp]
\centering
\includegraphics[width = 8cm]{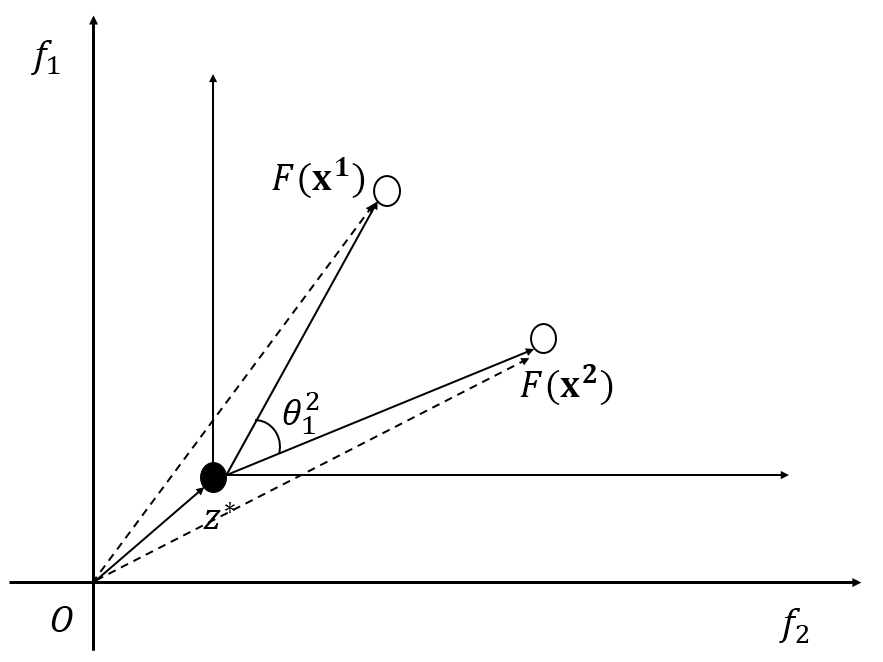}
\caption{Illustration of the angle between $\mathbf{x}^1$ and $\mathbf{x}^2$}
\label{fig:angle}
\end{figure*}

As shown in Fig. \ref{fig:angle}, given any two solutions $\mathbf{x}^1$ and $\mathbf{x}^2$, the angle between them in the objective space is $\theta_1^2$. Obviously, the angle between any two solutions is less than or equal to ${\pi}/{2}$, which means that the range of angle between any two solutions belongs in $[0, {\pi}/{2}].$

Given any two solutions $\mathbf{x}^1$ and $\mathbf{x}^2$, an threshold of angle $\theta$, a random number $r$ and a parameter $p_f$ ($\frac{\text{Number of Feasible Solutions}}{\text{Population Size}}$) which denotes the proportion of feasible solutions in the current population, the ACDP is defined as follows:
\begin{enumerate}
\item If $\mathbf{x}^1$ and $\mathbf{x}^2$ are both feasible, the one dominating the other is better.
\item If there is at lease one infeasible solution and $angle(\mathbf{x}^1,\mathbf{x}^2,z^*) \le \theta$, the one with a smaller constraint violation dominates the other.
\item When there is at least one infeasible solution and $angle(\mathbf{x}^1,\mathbf{x}^2,z^*) > \theta$, if $r < p_f$, the one dominating the other is better, otherwise, they are incomparable.
\end{enumerate}
\subsection{ACDP in the framework of MOEA/D}
\label{sec:3.2}

As we know, MOEA/D uses the value of decomposition function of a solution to update its neighbors. In order to use ACDP to handle constraints in the framework of MOEA/D, here we provide a version of ACDP which is suitable to the algorithm.

Given a subproblem $sp$ with the weight vector $\lambda$, for two solutions $\mathbf{x}^1$ and $\mathbf{x}^2$, their overall constraint violations are $\phi^1$ and $\phi^2$, and their decomposition values on the subproblem $sp$ are $g^{te}(\mathbf{x}^1|\lambda,z^{*})$ and $g^{te}(\mathbf{x}^2|\lambda,z^{*})$. The ACDP dominance $\preceq_{\theta}$ in the framework of MOEA/D is defined as follows:

\begin{eqnarray}
\label{equ:ACDP}
& \mathbf{x}^1 \preceq_{\theta}  \mathbf{x}^2\Leftrightarrow
& \begin{cases}
\textbf{Rule 1 }\text{if}\ \ \phi^1 = 0,\ \phi^2 = 0 : \ \ \ \ \\
\ \ \ \ \ \  g^{te}(\mathbf{x}^1|\lambda,z^{*})<g^{te}(\mathbf{x}^2|\lambda,z^{*});\\
\textbf{Rule 2 }\text{if}\ \ \phi^1 < \phi^2 :\ \ \\
\ \ \ \ \ \ angle(\mathbf{x}^1,\mathbf{x}^2,z^*) \le \theta; \\
\textbf{Rule 3 }\text{otherwise :} \\
\ \ \ \ \ \ angle(\mathbf{x}^1,\mathbf{x}^2,z^*) > \theta, r < p_f, \\
\ \ \ \ \ \ g^{te}(\mathbf{x}^1|\lambda,z^{*})<g^{te}(\mathbf{x}^2|\lambda,z^{*}).\\
\end{cases}
\end{eqnarray}
where $\theta$ is a threshold parameter, which is defined by users. In Eq. (\ref{equ:ACDP}), the constraint-handling method ACDP is equivalent to CDP \cite{996017} when $\theta \ge \frac{\pi}{2}$.

In Rule 1 of ACDP, when these two solutions are both feasible, the solution with a lower aggregation value dominates the other, which is similar to the first rule of CDP.

When at least one of $\mathbf{x}^1$ and $\mathbf{x}^2$ is infeasible, CDP only utilizes the constraint violations of these two solutions to compare, which is difficult to keep the diversity of the working population when most of solutions in the population are infeasible. Nevertheless, ACDP utilizes additional information to compare the two solutions, which includes the angle between the two compared solutions in the objective space and the proportion of feasible solutions in the current population ($p_f$). More details of ACDP in this situation are listed as follows:

\begin{itemize}
\item In Rule 2 of ACDP, if the angle between $\mathbf{x}^1$ and $\mathbf{x}^2$ in the objective space is smaller than the parameter $\theta$, ACDP considers that these two solutions are similar and compares them according to their constraint violations. Because these two solutions are similar, based on the framework of MOEA/D, they will be considered to relate to the same subproblem. In this situation, using the constraint violations to compare the two solutions will not cause the loss of the diversity.

\item In Rule 3 of ACDP, if the angle between $\mathbf{x}^1$ and $\mathbf{x}^2$ in the objective space is larger than the parameter $\theta$, ACDP considers that these two solutions are dissimilar, and the solution with a lower decomposition value will dominate the other with a probability of $p_f$. Some solutions with low aggregation values but large constraint violations will have a chance to be selected in the next generation, which can enhance the convergence of the working population effectively.    
\item The probability in Rule 3 of ACDP is set to be the proportion of feasible solutions in the current population. It keeps the balance of the exploration of the working population between infeasible regions and feasible regions. When $p_f$ is large, ACDP tends to explore infeasible regions. When $p_f$ is small, ACDP tends to explore feasible regions.

\end{itemize}
\subsection{Effectiveness of ACDP in MOEA/D}
\label{sec:3.3}
The evolutionary process of a CMOEA can be generally divided in three stages according to the status of the working population.

\begin{figure*}[htbp]
\centering
\begin{tabular}{cc}
\begin{minipage}[t]{0.5\linewidth}
\includegraphics[width = 5.5cm]{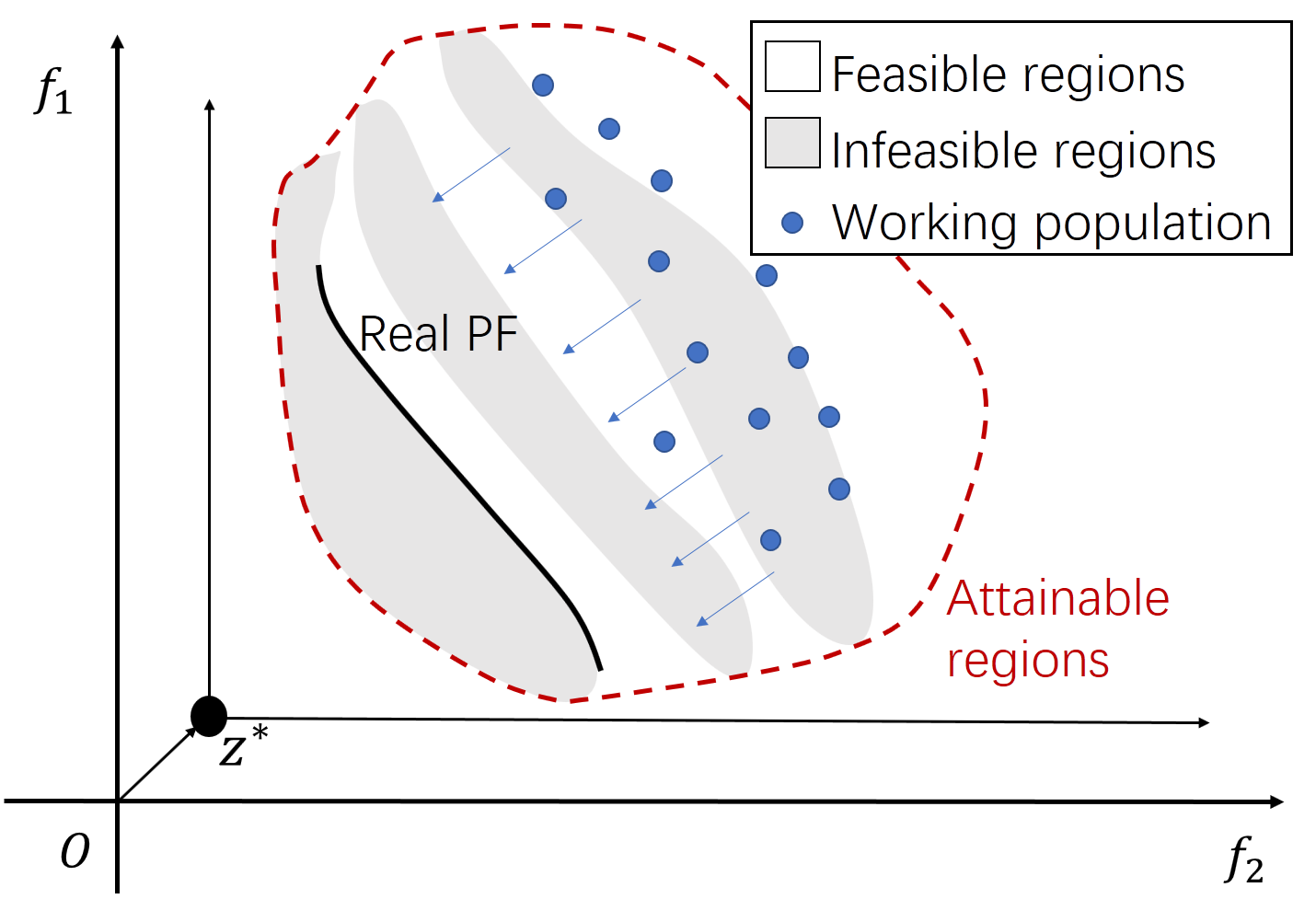}\\
\centering{\scriptsize{(a) Stage 1 of CDP}}
\end{minipage}
\begin{minipage}[t]{0.5\linewidth}
\includegraphics[width = 5.5cm]{CDP1.png}\\
\centering{\scriptsize{(b) Stage 1 of ACDP}}
\end{minipage}
\end{tabular}
\begin{tabular}{cc}
\begin{minipage}[t]{0.5\linewidth}
\includegraphics[width = 5.5cm]{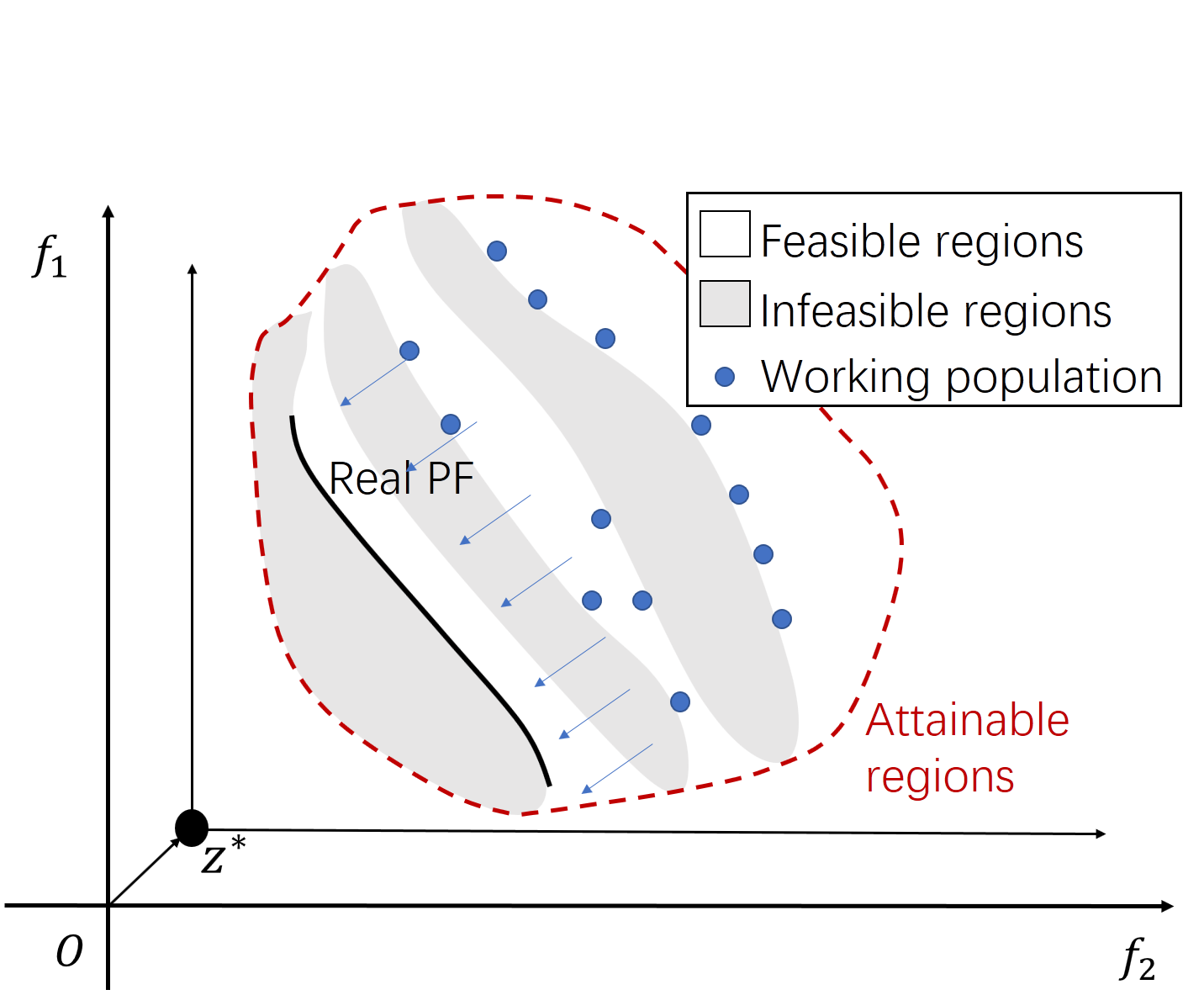}\\
\centering{\scriptsize{(c) Stage 2 of CDP}}
\end{minipage}
\begin{minipage}[t]{0.5\linewidth}
\includegraphics[width = 5.5cm]{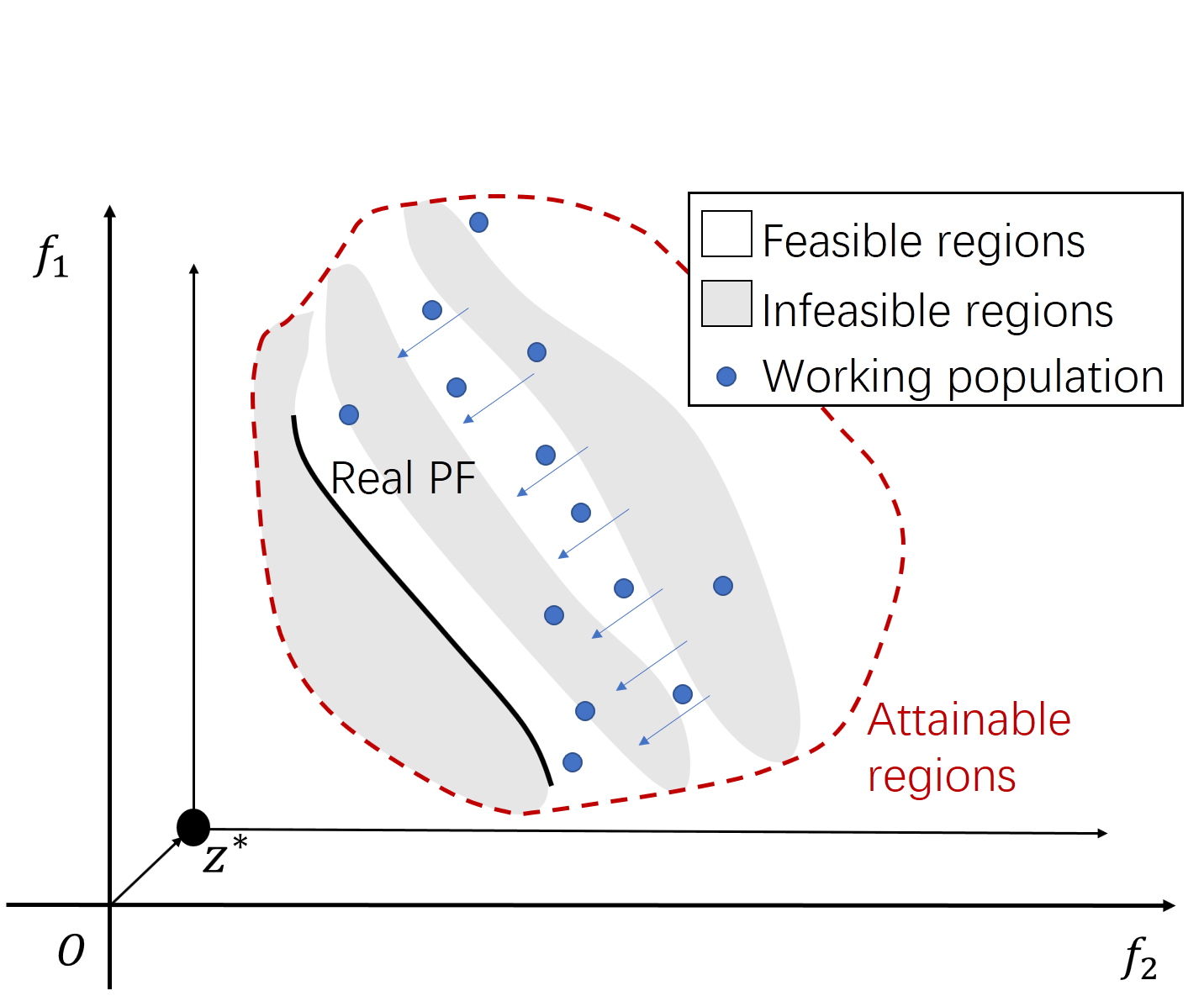}\\
\centering{\scriptsize{(d) Stage 2 of ACDP}}
\end{minipage}
\end{tabular}
\begin{tabular}{cc}
\begin{minipage}[t]{0.5\linewidth}
\includegraphics[width = 5.5cm]{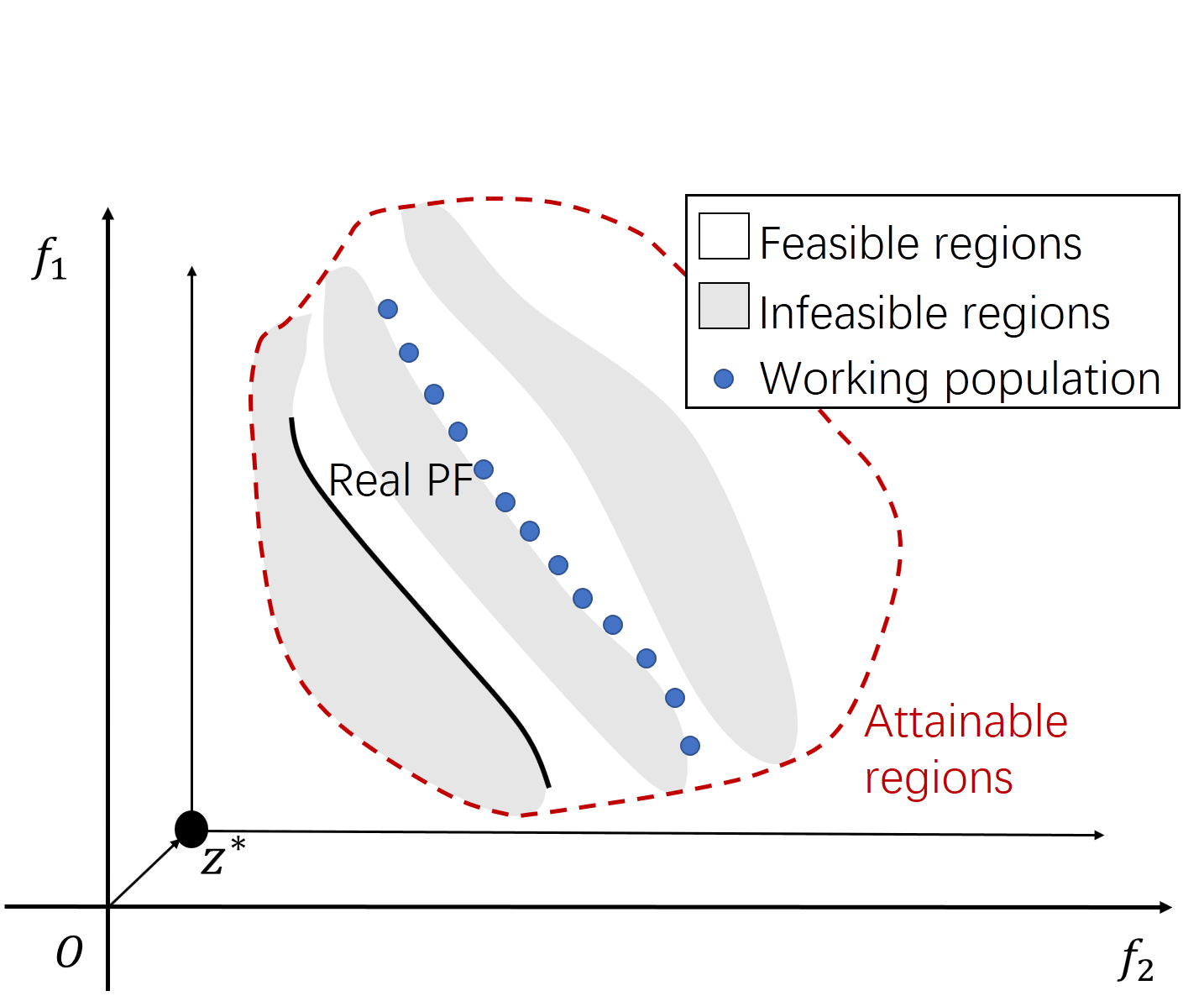}\\
\centering{\scriptsize{(e) Stage 3 of CDP}}
\end{minipage}
\begin{minipage}[t]{0.5\linewidth}
\includegraphics[width = 5.5cm]{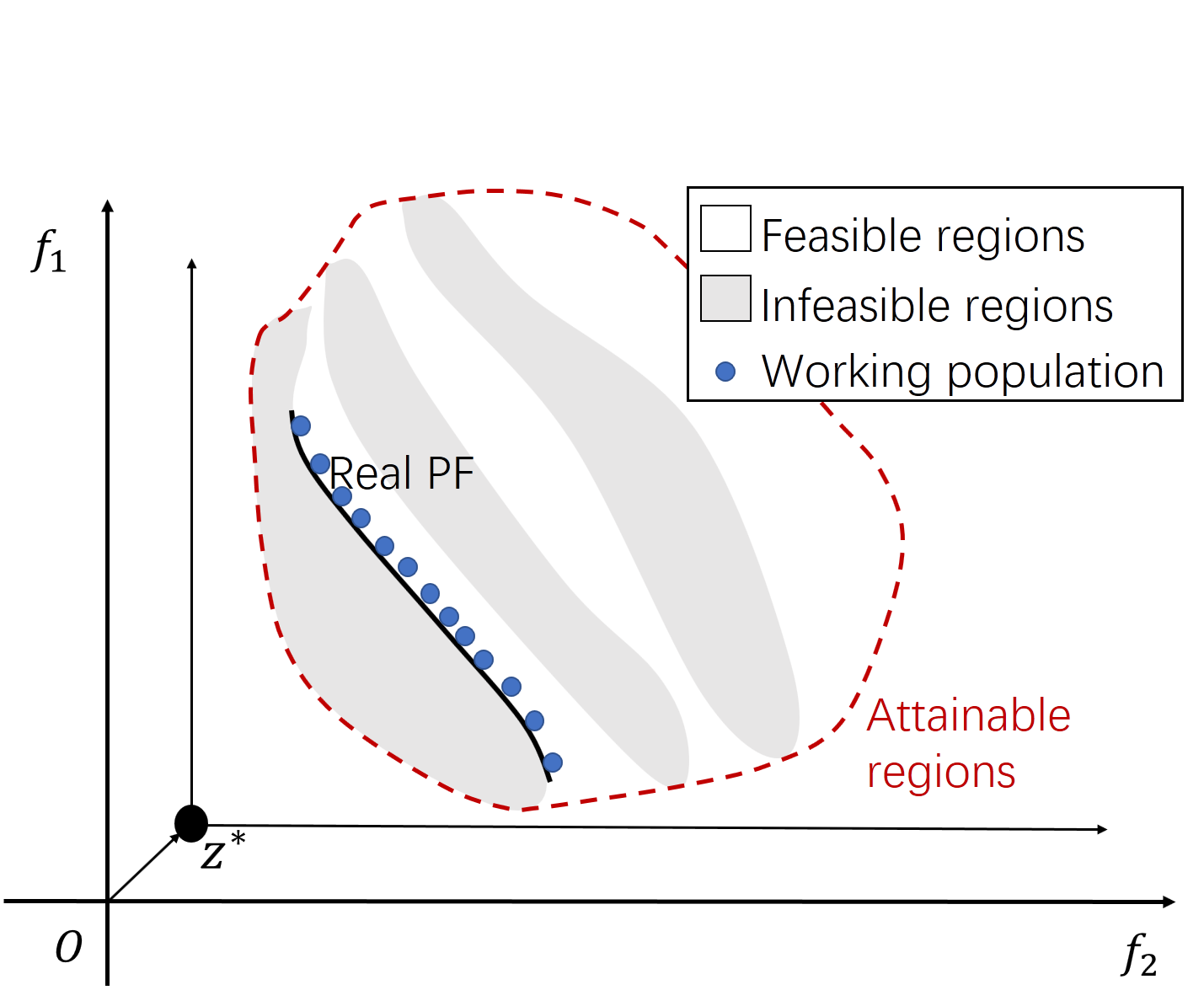}\\
\centering{\scriptsize{(f) Stage 3 of ACDP}}
\end{minipage}
\end{tabular}
\caption{Illustrations of the evolutionary process of MOEA/D with CDP and ACDP.}\label{Fig:CDP}
\end{figure*}

In the first stage, a population is generated randomly, and most of the individuals are far away from the real PF as shown in Fig. \ref{Fig:CDP} (a) and Fig. \ref{Fig:CDP} (b).

In the second stage, the working population begins to explore the search space. As shown in Fig.  \ref{Fig:CDP} (c), when using CDP in MOEA/D, the working population will be attracted to feasible regions and actually difficult to get across infeasible regions. As shown in Fig. \ref{Fig:CDP} (d), when ACDP is applied to MOEA/D, the working population can maintain the diversity by using angle information. Some individuals can enter infeasible regions, which can help the working population to go across infeasible regions effectively. Additionally, ACDP uses the information of the proportion of feasible solutions in the current population as the probability to select solutions, which can help to balance the search between feasible and infeasible regions.

In the third stage, the working population will converge to its near feasible regions. when using CDP, the population is trapped into local optimum, because of the difficulty to get across infeasible regions in the second stage, as shown in Fig. \ref{Fig:CDP} (e).  Conversely, when using ACDP, the working population can converge to the real PF more completely as shown in Fig. \ref{Fig:CDP} (f), because the population can keep the diversity and explore infeasible regions in the second stage.

\subsection{The Parameter Setting of Theta}
\label{sec:3.4}
 In the early stage of evolutionary process, the population is commonly far away from real PF. To prevent the population from being trapped into local optimum, the value of $\theta$ should be set small to maintain the diversity. In the later stage of evolutionary process, the convergence to the feasible regions should be emphasized, then the value of $\theta$ should become larger. Based on the above discussions, the threshold $\theta(k)$ should be dynamically increased with the generation counter $k$ increasing. A method of setting $\theta(k)$ is proposed as follows:

\begin{eqnarray}\label{equ:theta}
&\theta(k)=
& \begin{cases}
\theta_0\left(1+\frac{k}{T_{max}}\right)^{cp}, 1\le k \le T_c\\
\frac{\pi}{2}\quad \quad \quad \quad \quad \quad ,T_c < k\le T_{max}
& \end{cases}
\end{eqnarray}
where $\theta_0$ is an initial threshold value which is set as ${\pi}/{2N}$, $N$ is the size of population and $T_{max}$ is the maximum generation. $\alpha$ is a parameter, which is set as 0.8. $T_c=\alpha T_{max}$ is the termination generation to control $\theta$. Parameter $cp$ is initialized to ${\log (N)}/{\log (1+\alpha)}$.

\begin{figure*}[htbp]
\centering
\includegraphics[width = 8cm]{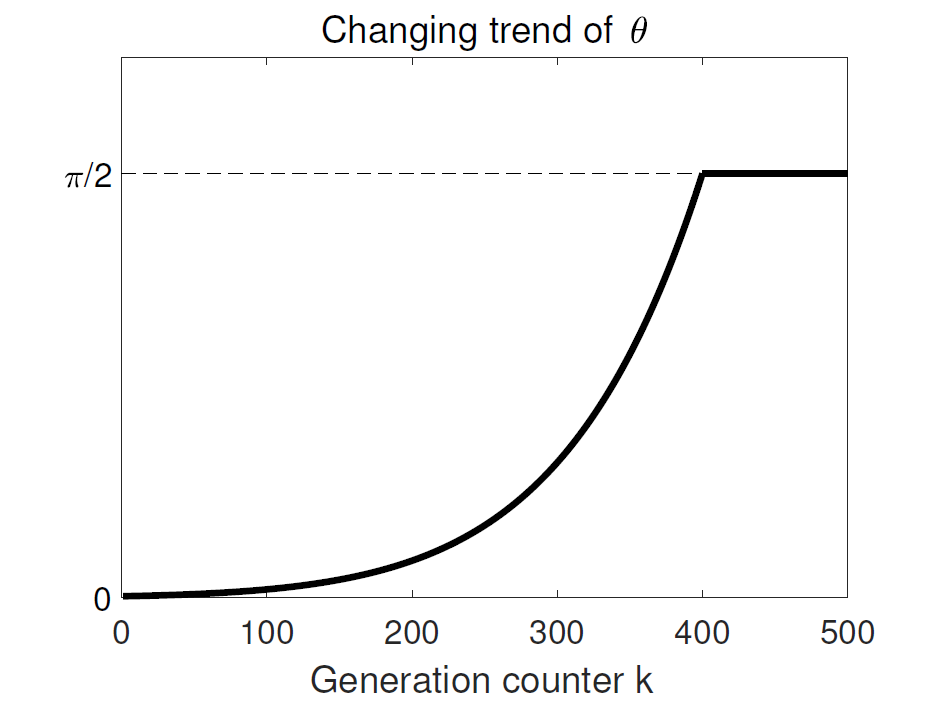}
\caption{The changing trend of $\theta$.}
\label{fig:theta}
\end{figure*}

As we know, the uniform weight vectors defined in Eq. (\ref{equ:weight}) decide that the maximum angle between two vectors is ${\pi}/{2}$, and the average angle between two adjacent vectors is ${\pi}/{2N}$, where $N$ is the size of population. Then, $\theta_0$ is initially set as ${\pi}/{2N}$. According to Eq. (\ref{equ:theta}), when the generation counter $k$ reaches $T_c$, the value of $\theta(k)$ is ${\pi}/{2}$, and keeps constant afterwards.
As shown in Fig. \ref{fig:theta}, we assume that the population size $N$ and the maximum generation $T_{max}$ are set as 300 and 500, respectively. Meanwhile $\alpha$ is set as 0.8. We can find that $\theta_0={\pi}/{2N}$. In the early stage of evolutionary process, $\theta(k)$ increases continuously but slowly. It benefits the population to maintain diversity. When $k$ gets more and more closed to $T_c$, $\theta(k)$ rises faster, which helps to accelerate the convergence speed to the feasible regions.
When $k$ reaches $T_c$, $\theta(k)$ is equal to ${\pi}/{2}$, ACDP is transformed into CDP.

\subsection{ACDP embedded in MOEA/D}
\label{sec:3.5}
The proposed MOEA/D-ACDP integrates the general framework of MOEA/D and the angle-based constrained dominance principle.

\begin{algorithm}
    \KwIn{\\
    $N$: the number of subproblems.\\
    $T_{max}$: the maximal generation.\\
    $N$ weight vectors: $\mathbf{\lambda}^1,\ldots,\mathbf{\lambda}^N$.\\
    $T$: the size of the neighborhood.\\
    $\delta$: the selecting probability from neighbors.\\
    $n_r$: the maximal number of solutions replaced by a child.\\
    $\theta_0$, $\alpha$: the parameters of ACDP method.
    }
    \KwOut{$NS:$ a set of feasible non-dominated solutions}

    Decompose a CMOP into $N$ subproblems associated with $\mathbf{\lambda}^1,\ldots,\mathbf{\lambda}^N$.\\
    Generate an initial population $P=\{\mathbf{x}^1, \ldots, \mathbf{x}^N \}$.\\
    Initialize $cp$ to be $\frac{\log (N)}{\log (1+\alpha)}$.\\
    Initialize the ideal point $z^*=(z_1,\ldots,z_m)$.\\
    For each $i = 1, \dots, N$, set $B(i) = \{i_1,\dots,i_T\}$, where $\mathbf{\lambda}^{i_1},\dots,\mathbf{\lambda}^{i_T}$ are the $T$ closest weight vectors to $\mathbf{\lambda}^i$.\\
    \For{$k \leftarrow 1$ \KwTo $T_{max}$}{
    \eIf{$k\le \alpha T_{max}$}{
    Set $\theta(k)$ according to $\theta(k)=\theta_0(1+\frac{k}{T_{max}})^{cp}$.\\
    }{Set $\theta(k)$ to be equal to $\frac{\pi}{2}$}
    Update $pf$ in the current generation.\\
    Generate a random permutation $rp$ from $\{1,\ldots,N\}$.\\
    \For{$i \leftarrow 1$ \KwTo $N$}{
    Generate a random number $r\in[0,1]$.\\
    $j = rp(i)$.\\
    \eIf{$r < \delta$}{
    $S = B(j)$
    }{
    $ S = \{1,\ldots,N\}$
    }
    Generate $\mathbf{y}^j$ through DE and polynomial mutation operators.\\

    Update the current ideal point.\\
    Set $c = 0$.\\
    \While{$c \neq n_r$ \rm{and} $S \neq \varnothing$}{
     select an index $j$ from $S$ randomly, $S = S \backslash\{j\}$.\\
     $result$ = $UpdateSubproblems$($\mathbf{x}^j$, $\mathbf{y}^j$, $\theta(k)$, $pf$)\\
     \lIf{$result == true$}{
     $c = c+1$}

    }

    }

    $NS$ = NondominatedSelect($NS  \bigcup P$)
    }
\caption{MOEA/D-ACDP}
\label{alg:moead-ACDP}
\end{algorithm}
The psuecode of MOEA/D-ACDP is listed in Algorithm \ref{alg:moead-ACDP}. Lines 1-5 initialize some parameters in MOEA/D-ACDP. First, a CMOP is decomposed into $N$ subproblems which are associated with weight vectors $\lambda^1,\ldots, \lambda^N$. Then the population $P$, the initial increasing factor $cp$, the ideal point $z^{*}$ and the neighbor indexes $B(i)$ are initialized. Lines 7-11 update the angle threshold value $\theta(k)$. Line 12 updates the proportion of feasible solutions in the current population $p_f$. Lines 13-23 generate a set of new solutions and update the ideal point $z^{*}$. To be more specific, lines 14-21 determine the set of neighboring solutions that may be updated by a newly generated solution $\mathbf{y}^j$. In line 22, the differential evolution (DE) crossover operator is adopted to generate a new solution $\mathbf{y}^j$. Meanwhile, $\mathbf{y}^j$ is further mutated by the polynomial mutation operator. The ideal point $z^{*}$ is updated in line 23. Lines 24-39 update subproblems. In line 27, the subproblems are updated based on the ACDP approach whose detailed psuecode is listed in Algorithm \ref{alg:subproblem}. At the end of each generation, non-dominated solutions ($NS$) in the population are selected to update the external archive based on non-dominated sorting in line 31.

\begin{algorithm}
\Fn{result = UpdateSubproblems($\mathbf{x}^j$, $\mathbf{y}^j$, $\theta(k)$, $pf$)}{
    $result = false$

    \eIf{$\phi(\mathbf{y}^j) ==0$ \rm{and} $\phi(\mathbf{x}^j) ==0$}{
        \If{$g^{te}(\mathbf{y}^i|\lambda^j,z^{*}) \leq g^{te}(\mathbf{x}^j|\lambda^j,z^{*})$}{
            $\mathbf{x}^j$ = $\mathbf{y}^j$\\
            $result = ture$
        }

   }
{ 
        \eIf{ $angle(\mathbf{F}(\mathbf{y}^j),\mathbf{F}(\mathbf{x}^j),z^{*})< \theta(k)$}{
         \If{$\phi(\mathbf{y}^j) < \phi(\mathbf{x}^j)$}{
            $\mathbf{x}^j$ = $\mathbf{y}^j$\\
            $result = ture$}

        }{        \If{$rand()<p_f$}{
         \If{$g^{te}(\mathbf{y}^i|\lambda^j,z^{*}) \leq g^{te}(\mathbf{x}^j|\lambda^j,z^{*})$}{
            $\mathbf{x}^j$ = $\mathbf{y}^j$\\
            $result = ture$
        }
              }}

}

    \Return $result$
}
\caption{Subproblem Update}
\label{alg:subproblem}
\end{algorithm}
In Algorithm \ref{alg:subproblem}, the algorithm updates a subproblem in terms of Eq. (\ref{equ:ACDP}). Lines 3-7 denote that when two feasible solutions $\mathbf{x}^j$ and $\mathbf{y}^j$ are compared, the one with a better aggregation value is selected. Lines 9-12 denote that when at least one of two solutions $\mathbf{x}^j$ and $\mathbf{y}^j$ is infeasible, if the angle between them in the objective space is lower than $\theta$, the solution with a lower constraint violation is selected. Lines 13-17 denote that when at least one of two solutions $\mathbf{x}^j$ and $\mathbf{y}^j$ is infeasible, if the angle between them in the objective space is larger than $\theta$, the solution with a lower aggregation value will be selected with a probability of $p_f$.

\section{Experimental Study}
\label{sec:4}
\subsection{Test Instances LIR-CMOPs}
\label{sec:4.1}
To evaluate the performance of the proposed MOEA/D-ACDP, 14 constrained multi-objective test problems with large infeasible regions in the objective space are used \cite{fan2016difficulty, Fan2017A}.

\begin{figure*}[htbp]
\begin{tabular}{cc}
\begin{minipage}[t]{0.33\linewidth}
\includegraphics[width = 4cm]{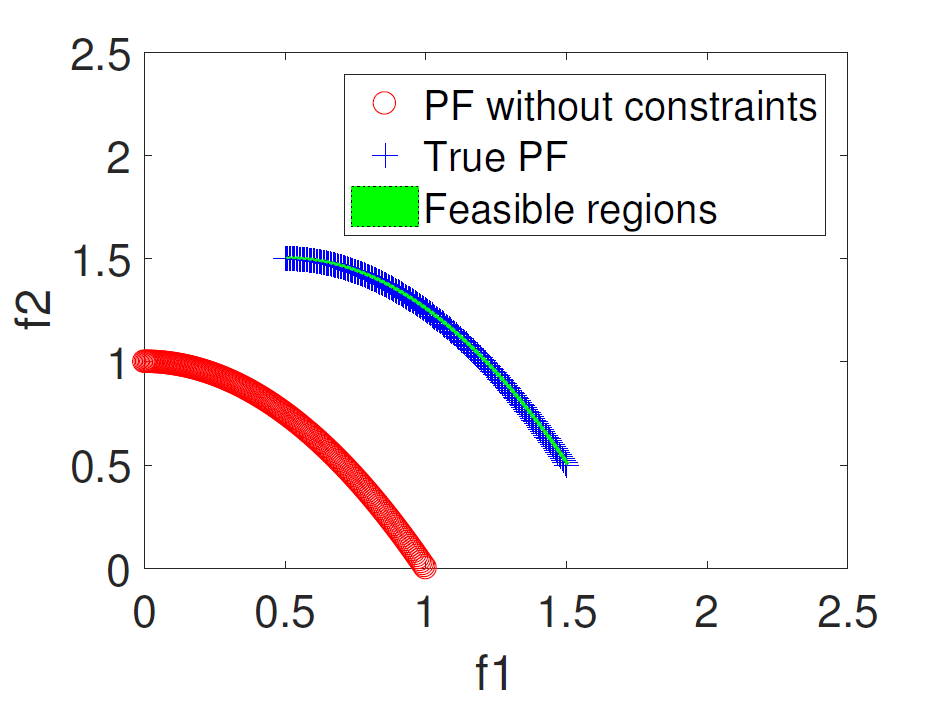}\\
\centering{\scriptsize{(a) LIR-CMOP1}}
\end{minipage}
\begin{minipage}[t]{0.33\linewidth}
\includegraphics[width = 4cm]{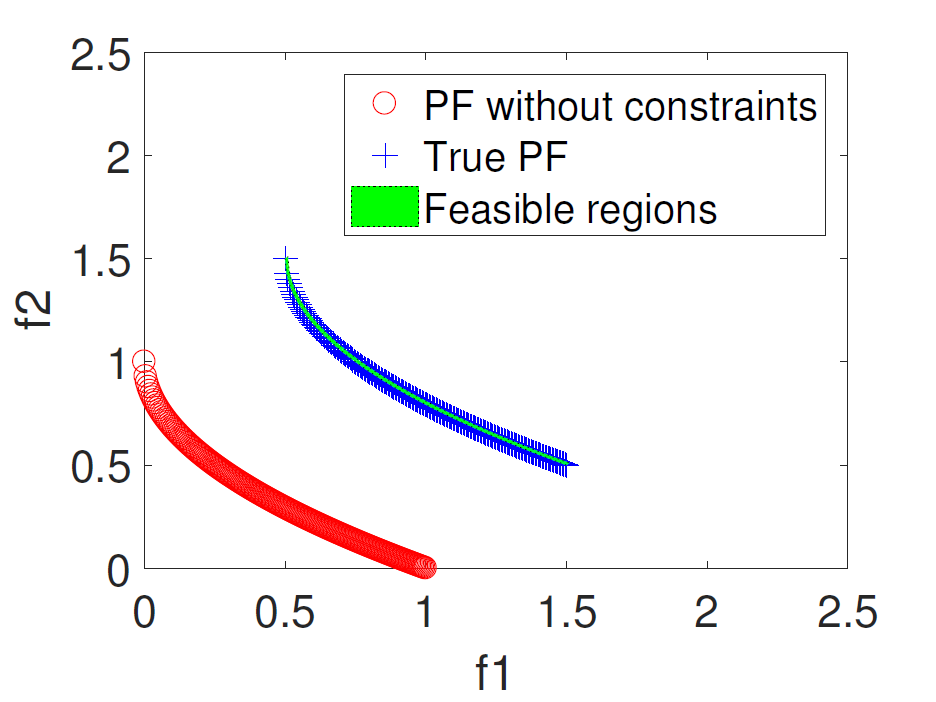}\\
\centering{\scriptsize{(b) LIR-CMOP2}}
\end{minipage}
\begin{minipage}[t]{0.33\linewidth}
\includegraphics[width = 4cm]{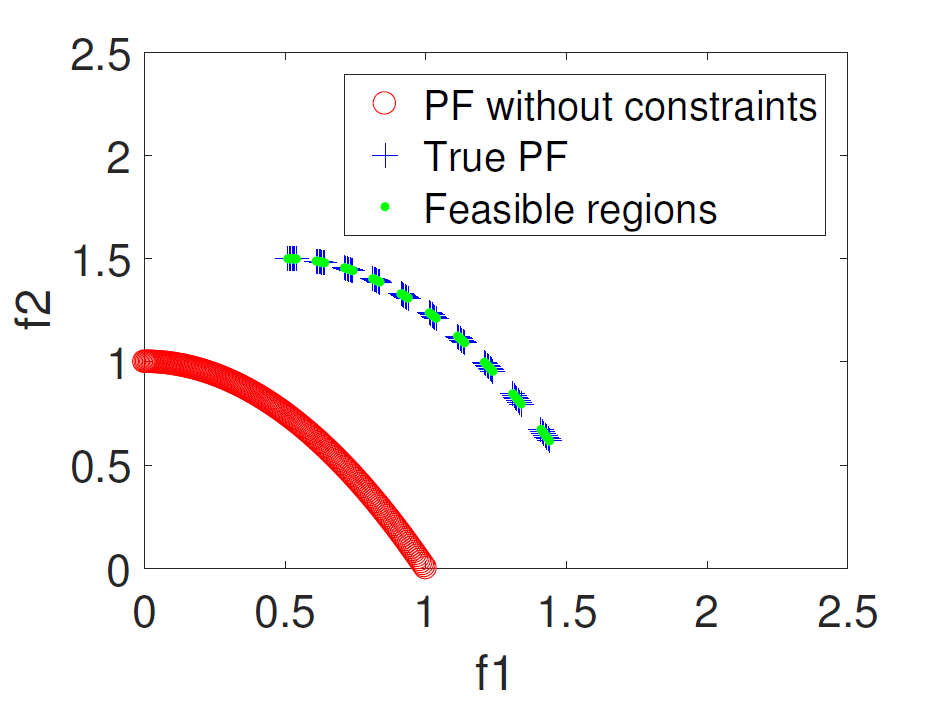}\\
\centering{\scriptsize{(c) LIR-CMOP3}}
\end{minipage}
\end{tabular}

\begin{tabular}{cc}
\begin{minipage}[t]{0.33\linewidth}
\includegraphics[width = 4cm]{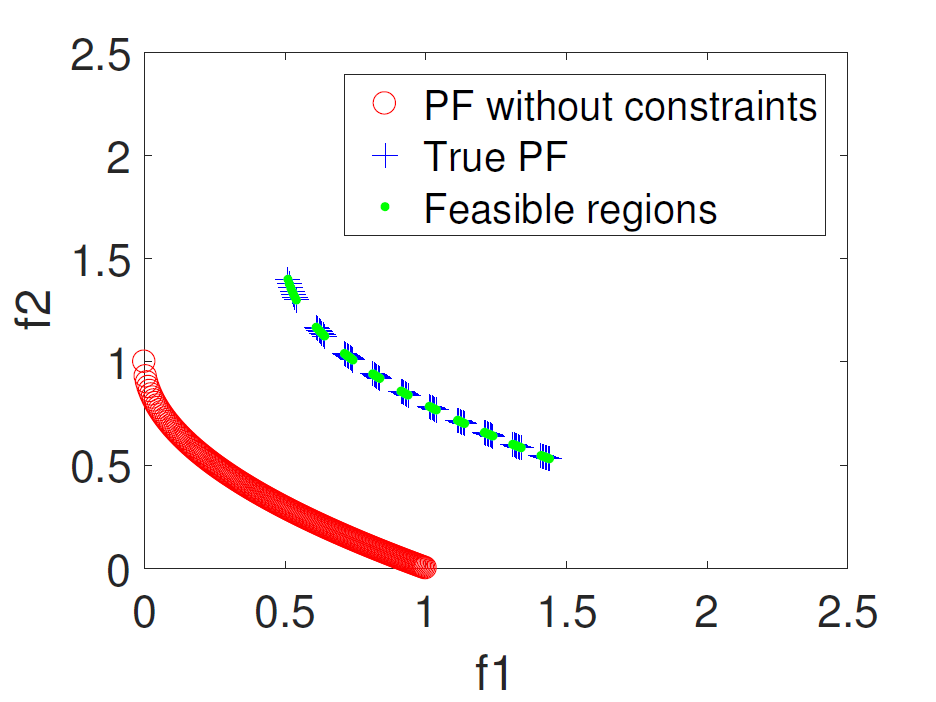}\\
\centering{\scriptsize{(d) LIR-CMOP4}}
\end{minipage}
\begin{minipage}[t]{0.33\linewidth}
\includegraphics[width = 4cm]{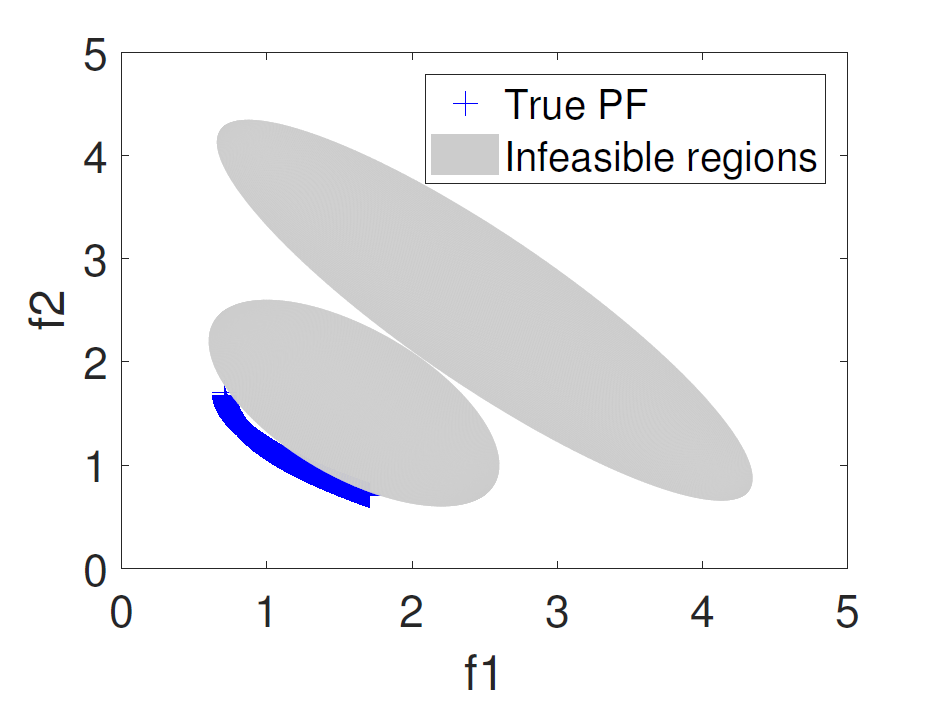}\\
\centering{\scriptsize{(e) LIR-CMOP5}}
\end{minipage}
\begin{minipage}[t]{0.33\linewidth}
\includegraphics[width = 4cm]{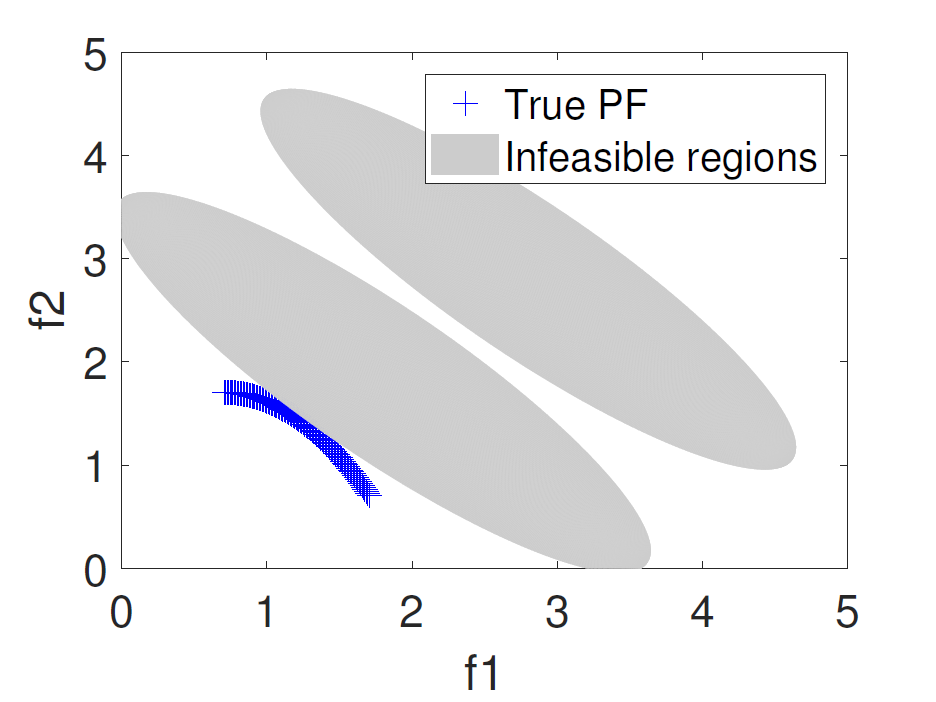}\\
\centering{\scriptsize{(f) LIR-CMOP6}}
\end{minipage}
\end{tabular}

\begin{tabular}{cc}
\begin{minipage}[t]{0.33\linewidth}
\includegraphics[width = 4cm]{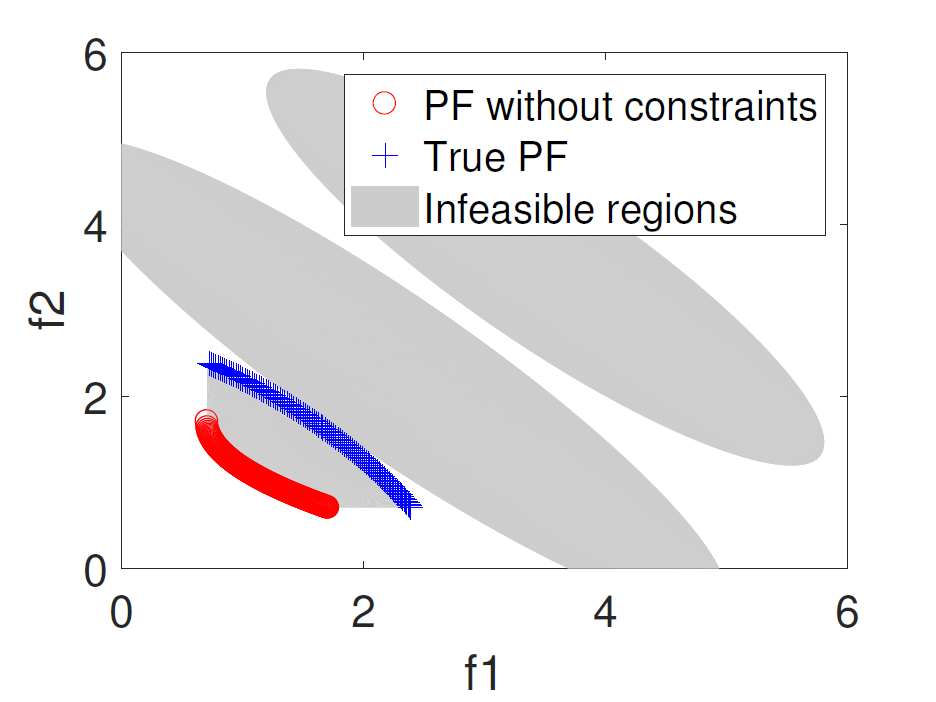}\\
\centering{\scriptsize{(g) LIR-CMOP7}}
\end{minipage}
\begin{minipage}[t]{0.33\linewidth}
\includegraphics[width = 4cm]{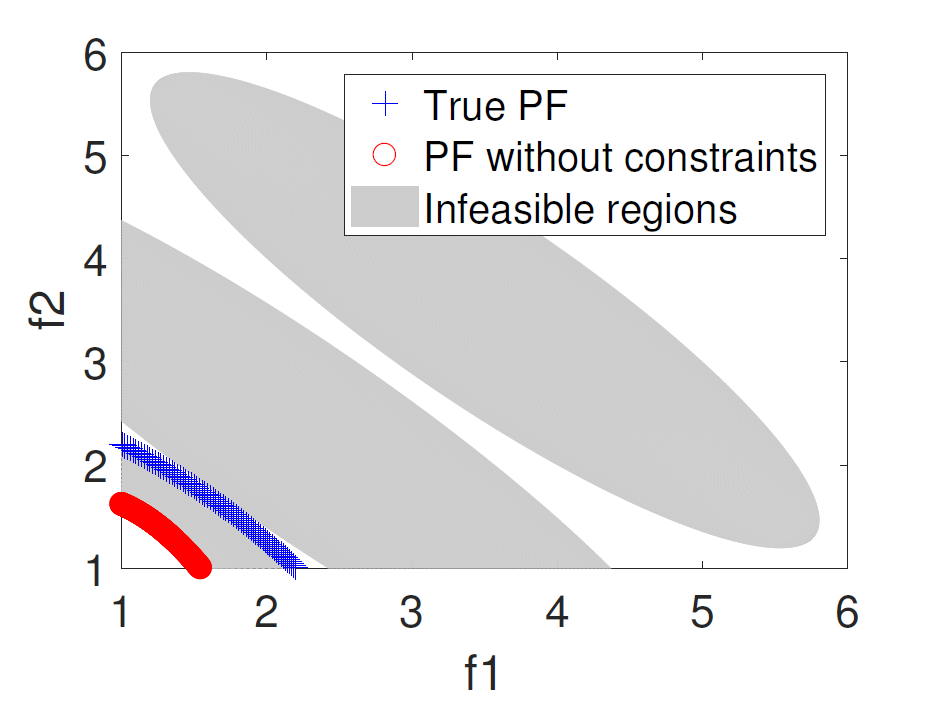}\\
\centering{\scriptsize{(h) LIR-CMOP8}}
\end{minipage}
\begin{minipage}[t]{0.33\linewidth}
\includegraphics[width = 4cm]{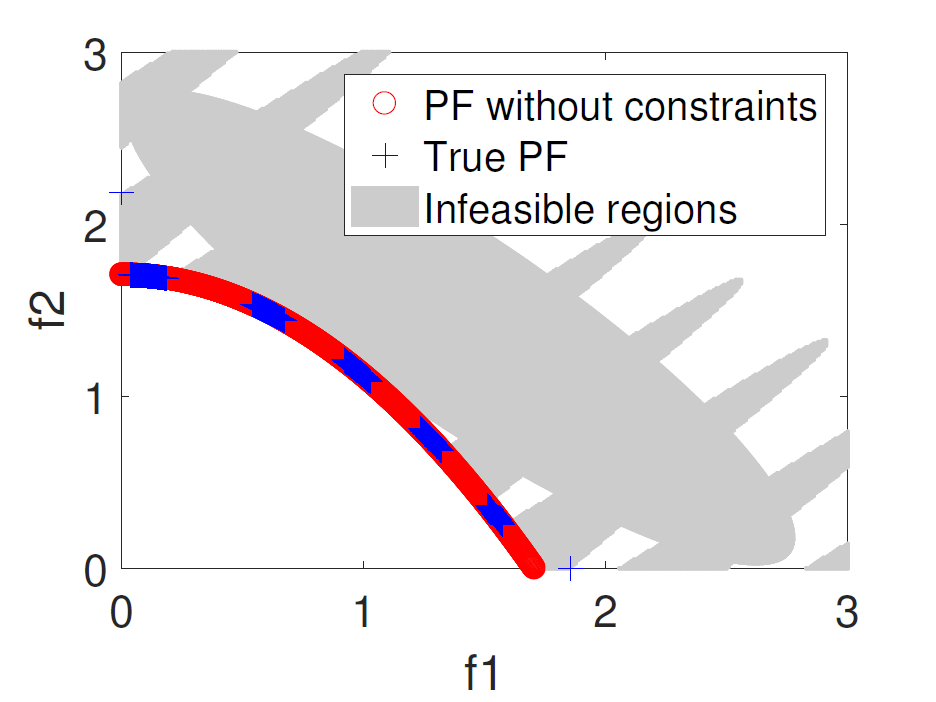}\\
\centering{\scriptsize{(i) LIR-CMOP9}}
\end{minipage}
\end{tabular}

\begin{tabular}{cc}
\begin{minipage}[t]{0.33\linewidth}
\includegraphics[width = 4cm]{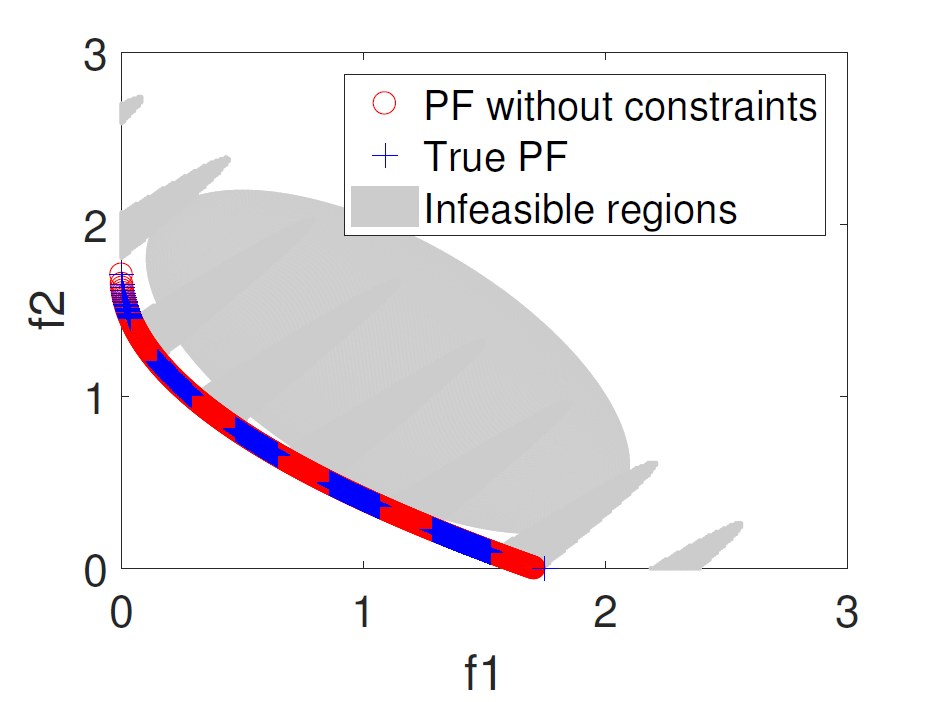}\\
\centering{\scriptsize{(j) LIR-CMOP10}}
\end{minipage}
\begin{minipage}[t]{0.33\linewidth}
\includegraphics[width = 4cm]{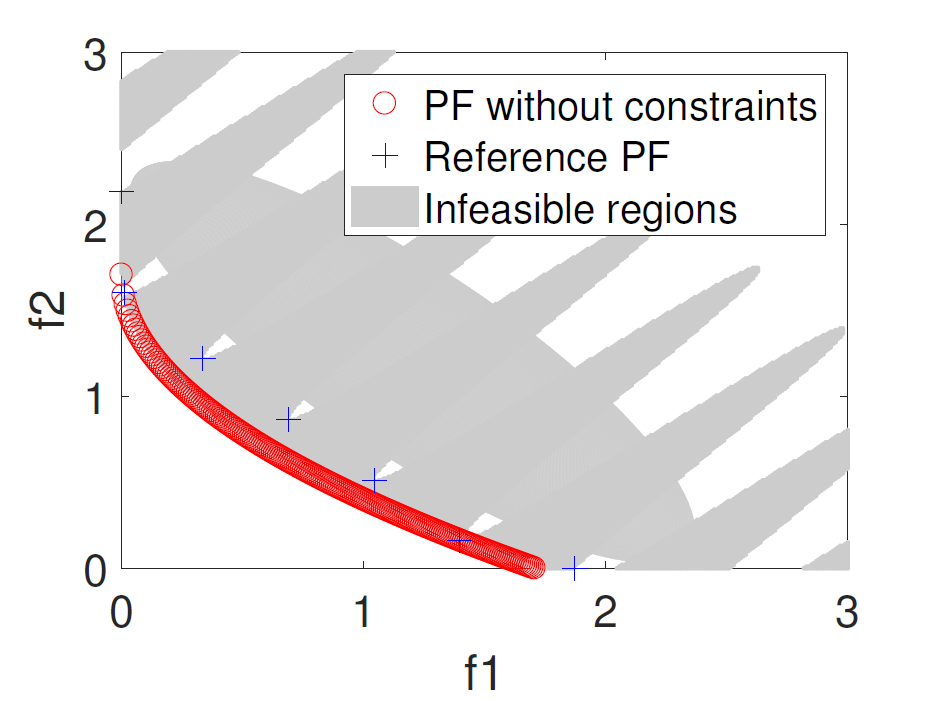}\\
\centering{\scriptsize{(k) LIR-CMOP11}}
\end{minipage}
\begin{minipage}[t]{0.33\linewidth}
\includegraphics[width = 4cm]{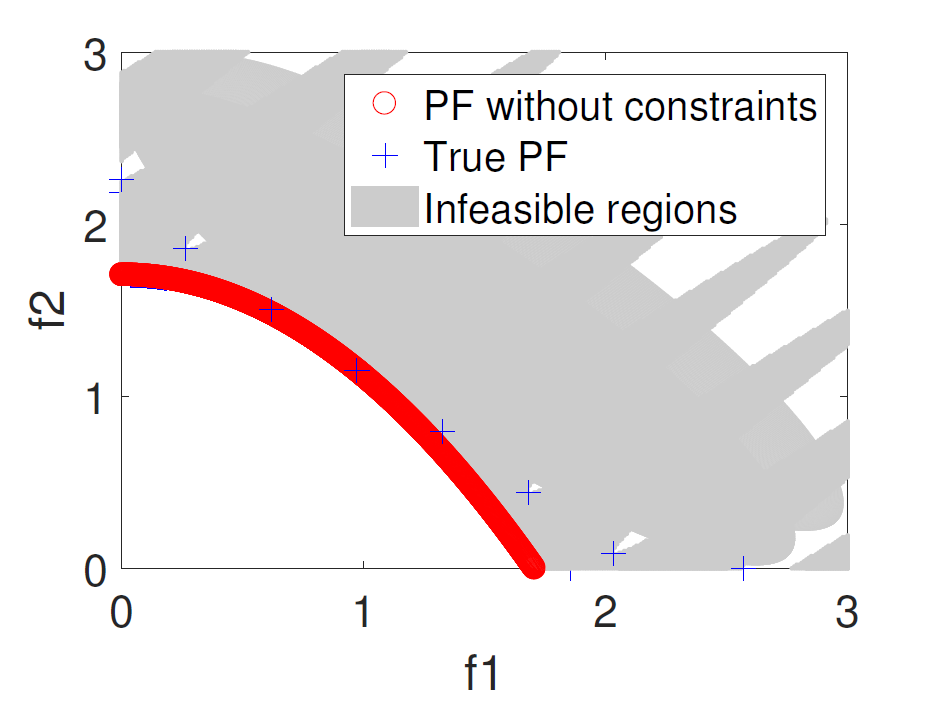}\\
\centering{\scriptsize{(l) LIR-CMOP12}}
\end{minipage}
\end{tabular}
\caption{Illustrations of the feasible and infeasible regions of LIR-CMOP1-12.} \label{Fig:lir-cmop-a}
\end{figure*}

\begin{figure*}[htbp]
\begin{tabular}{cc}
\begin{minipage}[t]{0.33\linewidth}
\includegraphics[width = 4cm]{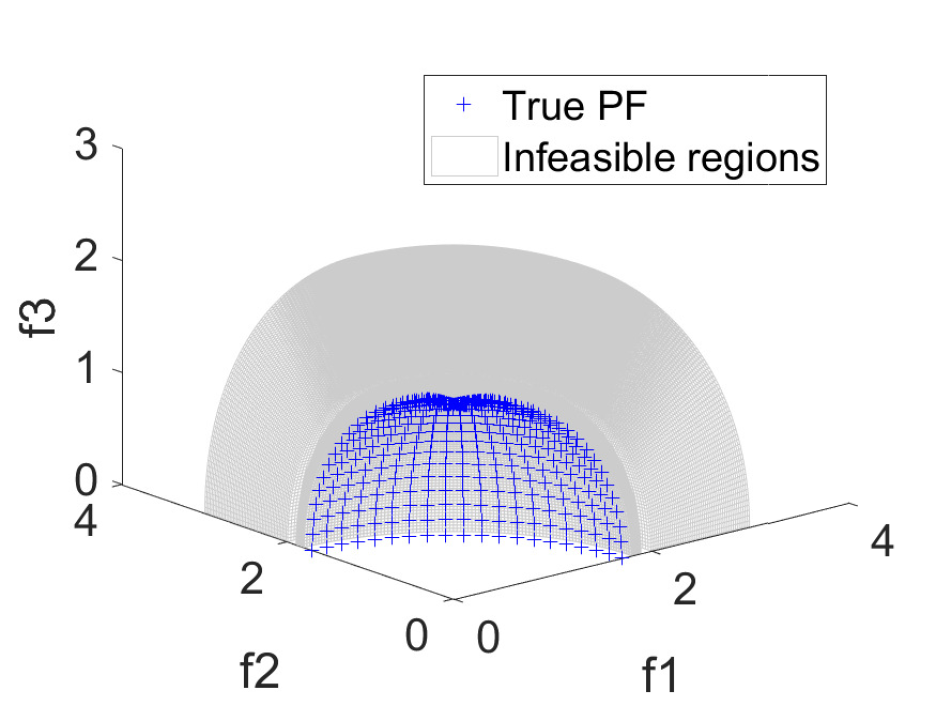}\\
\centering{\scriptsize{(a) LIR-CMOP13}}
\end{minipage}
\begin{minipage}[t]{0.33\linewidth}
\includegraphics[width = 4cm]{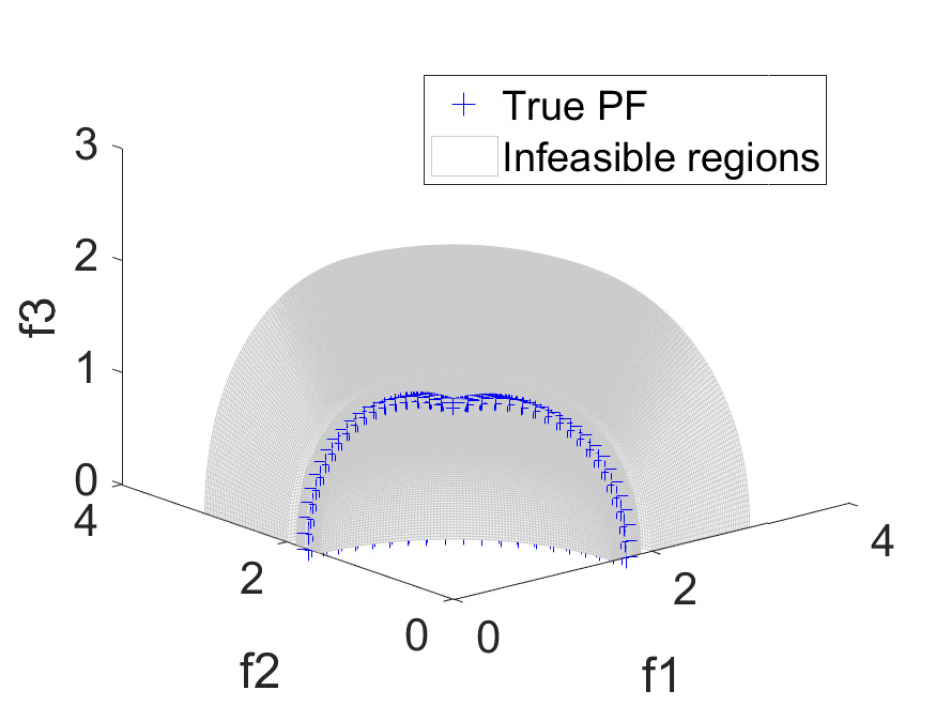}\\
\centering{\scriptsize{(b) LIR-CMOP14}}
\end{minipage}
\end{tabular}
\caption{Illustrations of the infeasible regions of LIR-CMOP13-14.} \label{Fig:lir-cmop-b}
\end{figure*}

The general characteristic of LIR-CMOPs is that their real PFs are blocked by a number of large infeasible regions, and thus hard to be found during an evolutionary process. Their constraint functions are comprised of controllable shape functions and distance functions \cite{Huband:2006hi}. More specifically, the shape functions are used to turn the PF shapes as convex or concave, while the distance functions are adopted to adjust the convergence difficulty for CMOEAs. Fig. \ref{Fig:lir-cmop-a} and Fig. \ref{Fig:lir-cmop-b} plot the feasible regions of LIR-CMOPs with two or three objectives, respectively.

\subsection{Real-world Engineering Optimization: I-beam}
\label{sec:4.2}
To evaluate the performance of MOEA/D-ACDP for solving real world optimization problems, an engineering optimization problem with two conflicting objectives is studied.

As defined in \cite{Osyczka1985Multicriteri}, an optimization problem of I-beam is a bi-objective constrained optimization problem which needs to minimize the following objectives simultaneously:\\
1. Cross section area of the beam;\\
2. Static deflection of the beam for the displacement under the force $P$.\\

\begin{figure*}[htbp]
\centering
\includegraphics[width = 8cm]{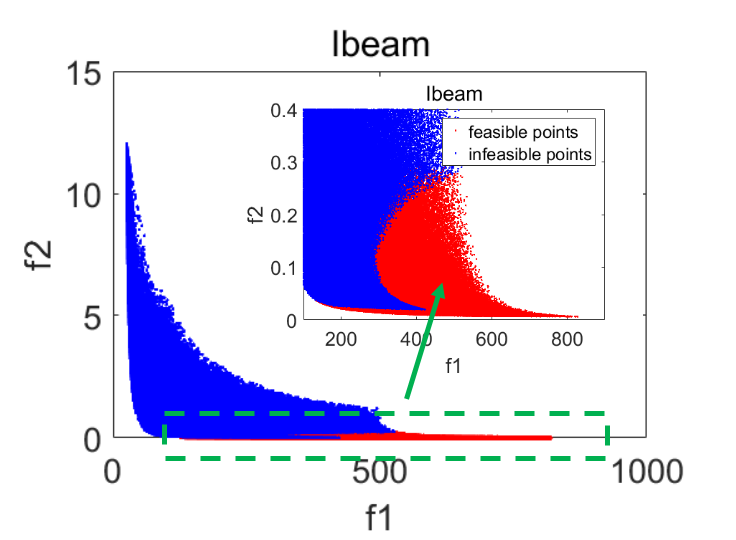}
\caption{The distribution of the I-Beam problem.}
\label{fig:sampling}
\end{figure*}

To study the landscape in the objective space of the I-beam optimization problem, 1,000,000 sampling solutions are generated, where 850,000 solutions are generated randomly, and the other 150,000 solutions are generated by MOEA/D-ACDP. In Fig. \ref{fig:sampling}, it is observed that there exist a few infeasible regions (the proportion of feasible solutions in all sampling solutions $p=0.5339$, which means that nearly a half of selected points are infeasible.) in the objective space for the I-beam optimization problem.

\subsection{Experimental Settings}
\label{sec:4.3}
To evaluate the performance of the proposed MOEA/D-ACDP, four popular CMOEAs (C-MOEA/D, MOEA/D-CDP, MOEA/D-Epsilon and MOEA/D-SR), with differential evolution (DE) crossover operator, are adopted and tested on LIR-CMOP1-14 and I-beam optimization problem. The detailed parameters are listed as follows:
\begin{enumerate}
\item Mutation probability $Pm = 1/n$ ($n$ is the number of decision variables) and its distribution index is set to 20. For DE operator, $CR = 1.0$, $f = 0.5$.
\item Population size: $N = 300$. Neighborhood size: $T = 30$.
\item Stopping condition: each algorithm runs for 30 times independently, and stops when 150,000 function evaluations are reached.
\item Probability of selecting individuals in the neighborhood: $\delta = 0.9$.
\item The maximal number of solutions replaced by a child: $nr = 2$.
\item Parameter setting in MOEA/D-ACDP: $\alpha = 0.8$ and $\theta_0 = {\pi }/{2N}$.
\item Parameter setting in MOEA/D-Epsilon: $T_c = 400$, $cp = 2$ and $\theta = 0.05 N$.
\item Parameter setting in MOEA/D-SR: $S_r = 0.01$.
\end{enumerate}

\subsection{Performance Metric}
\label{sec:4.4}
To measure the performance of MOEA/D-ACDP, C-MOEA/D, MOEA/D-CDP, MOEA/D-Epsilon and MOEA/D-SR, two widely used metrics inverted generation distance ($IGD$) \cite{Bosman2003The} and hypervolume ($HV$) \cite {Zitzler1999Multiobjective} are adopted as evaluation metrics. Their definitions are listed as follows.

\begin{itemize}
\item \textbf{Inverted Generational Distance} ($IGD$):
\end{itemize}
$IGD$ is a metric which evaluates the performance related to convergence and diversity simultaneously. Let $P^*$ be a set of uniformly distributed points in the ideal PF. Let $A$ denote an approximate PF achieved by a certain CMOEA.
The metric $IGD$ that represents average distance from $P^*$ to $A$ is defined as:
\begin{equation} \label{IGD metric}
\begin{cases}
IGD(P^*,A) = \dfrac{\sum \limits_{y^* \in P^*}d(y^*,A)}{| P^* |}\\
\\
d(y^*,A) = \min \limits_{y \in A} \{ \sqrt {\sum_{i = 1} ^m (y^{*}_{i} - y_i)^2} \}
\end{cases}
\end{equation}
In our experiment, for CMOPs with two objectives, 1000 points are sampled uniformly from the PF to constitute $P^*$.  For CMOPs with three objectives, 10000 points are sampled uniformly from the PF to constitute $P^*$.
A smaller $IGD$ represents a better performance regarding to both diversity and convergence.

\begin{itemize}
\item \textbf{Hypervolume} ($HV$):
\end{itemize}
$HV$ reflects the closeness between the non-dominated set achieved by a CMOEA and the representative PF. The larger $HV$ means that the corresponding non-dominated set is closer to the true PF.

\begin{equation}
HV(S)=VOL\left(\bigcup \limits_{\mathbf{x}\in S} [f_1(\mathbf{x}),z_1^r]\times ...[f_m(\mathbf{x}),z_m^r] \right) \\
\end{equation}
where $VOL(\cdot)$ is the Lebesgue measure, $\mathbf{z}^r=(z_1^r,...,z_m^r)^T$ is a reference point in the objective space. For the test instances LIR-CMOPs, the reference point is set as 1.4 times the nadir point of the real PF. The $HV$ with a larger value represents the better performance regarding to both diversity and convergence.
\\

As the real PF of the I-beam optimization problem is not known, $IGD$ metric can not be calculated. The experiment uses the $HV$ metric \cite {Zitzler1999Multiobjective} to measure the performance of these CMOEAs.
In the I-beam optimization case, the reference point is set as $z^r = [850,0.0615]^T$.

\subsection{Discussion of Experimental Results}
\label{sec:4.5}
\subsubsection{Performance Evaluation on LIR-CMOP Test Instances}
\label{sec:4.5.1}
The results of the $IGD$ values on LIR-CMOP1-14 achieved by five CMOEAs in 30 independent runs are shown in Table \ref{tab:lir-cmop-igd}.

 As discussed in Section \ref{sec:4}, LIR-CMOP1-14 all have large infeasible regions in their objective space. For LIR-CMOP3-14, MOEA/D-ACDP significantly outperforms the other four compared algorithms in terms of the $IGD$ metric.
 For LIR-CMOP1-2, MOEA/D-ACDP significantly outperforms C-MOEA/D, MOEA/D-CDP and MOEA/D-Epsilon.

The results of the $HV$ values on LIR-CMOP1-14 achieved by five CMOEAs in 30 independent runs are shown in Table \ref{tab:lir-cmop-hv}. For LIR-CMOP2-14, MOEA/D-ACDP significantly outperforms the compared algorithms in terms of the $HV$ metric. For the LIR-CMOP1, MOEA/D-ACDP significantly outperforms C-MOEA/D, MOEA/D-CDP and MOEA/D-Epsilon.

\begin{figure*}[htbp]
\centering
\begin{tabular}{cc}
\begin{minipage}[t]{0.25\linewidth}
\includegraphics[width = 3.3cm]{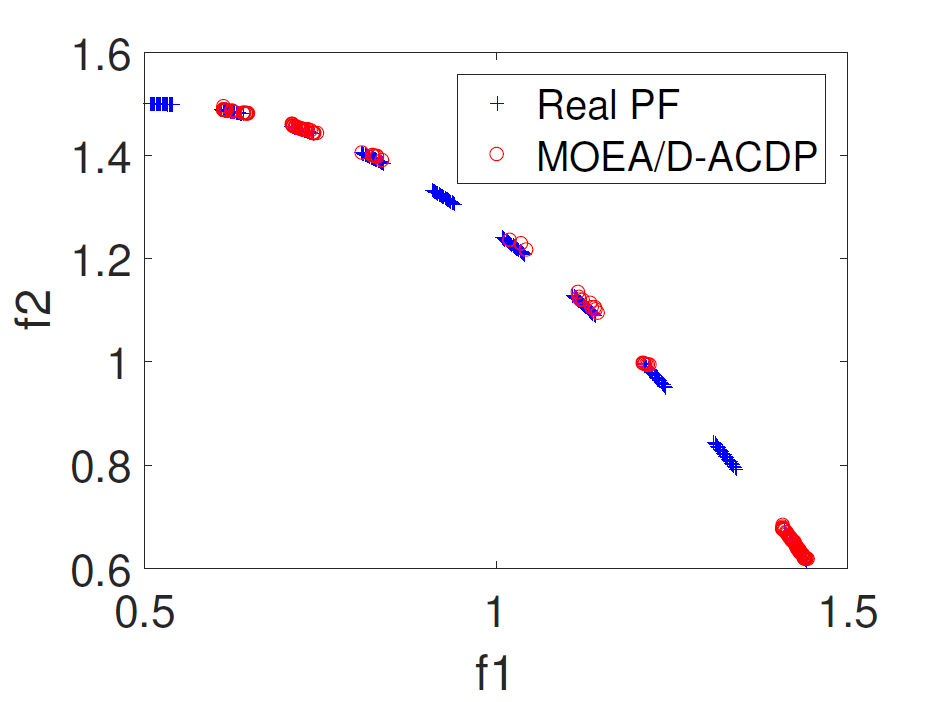}\\
\end{minipage}
\begin{minipage}[t]{0.25\linewidth}
\includegraphics[width = 3.3cm]{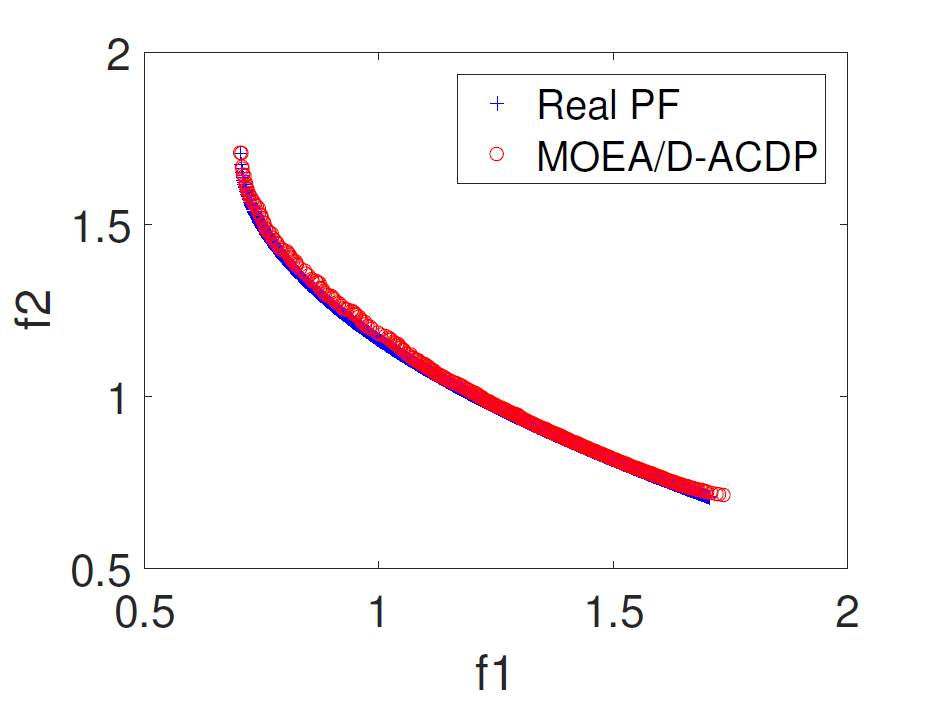}\\
\end{minipage}
\begin{minipage}[t]{0.25\linewidth}
\includegraphics[width = 3.3cm]{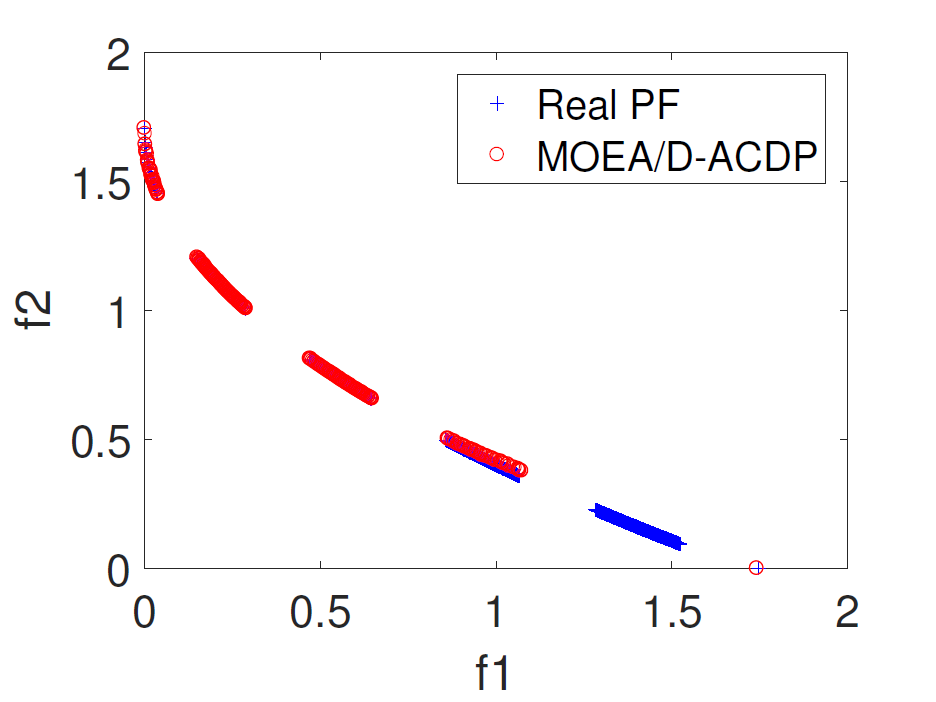}\\
\end{minipage}
\begin{minipage}[t]{0.25\linewidth}
\includegraphics[width = 3.3cm]{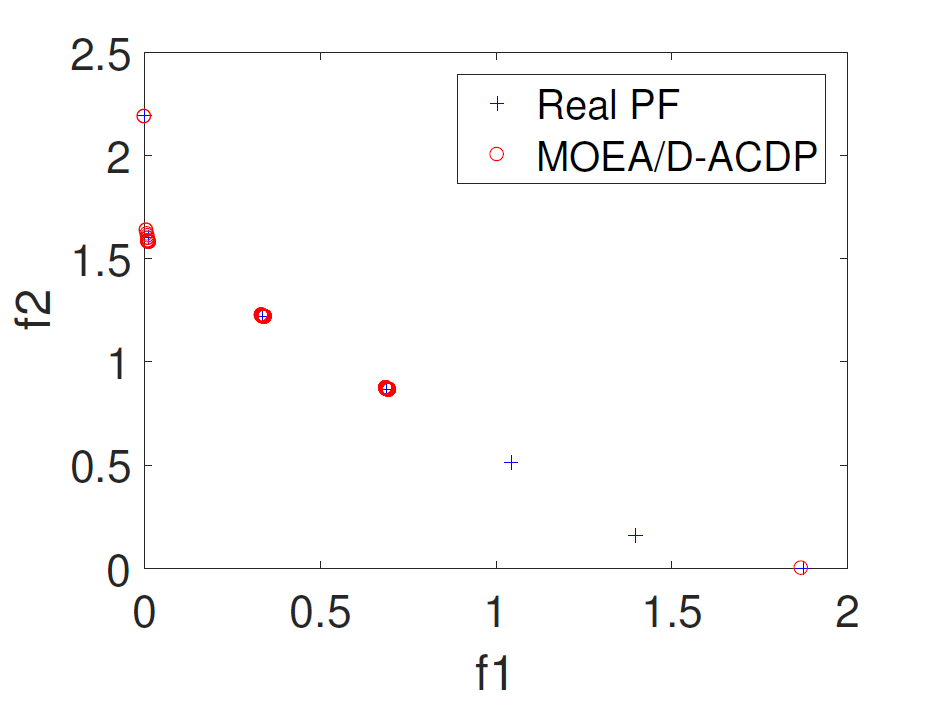}\\
\end{minipage}
\end{tabular}

\begin{tabular}{cc}
\begin{minipage}[t]{0.25\linewidth}
\includegraphics[width = 3.3cm]{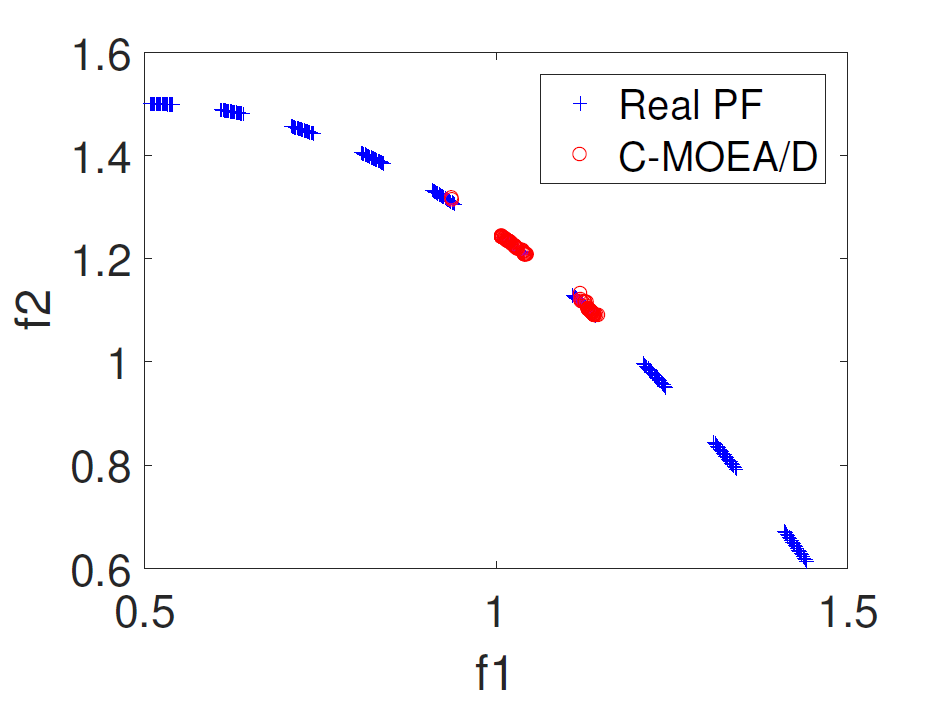}\\
\end{minipage}
\begin{minipage}[t]{0.25\linewidth}
\includegraphics[width = 3.3cm]{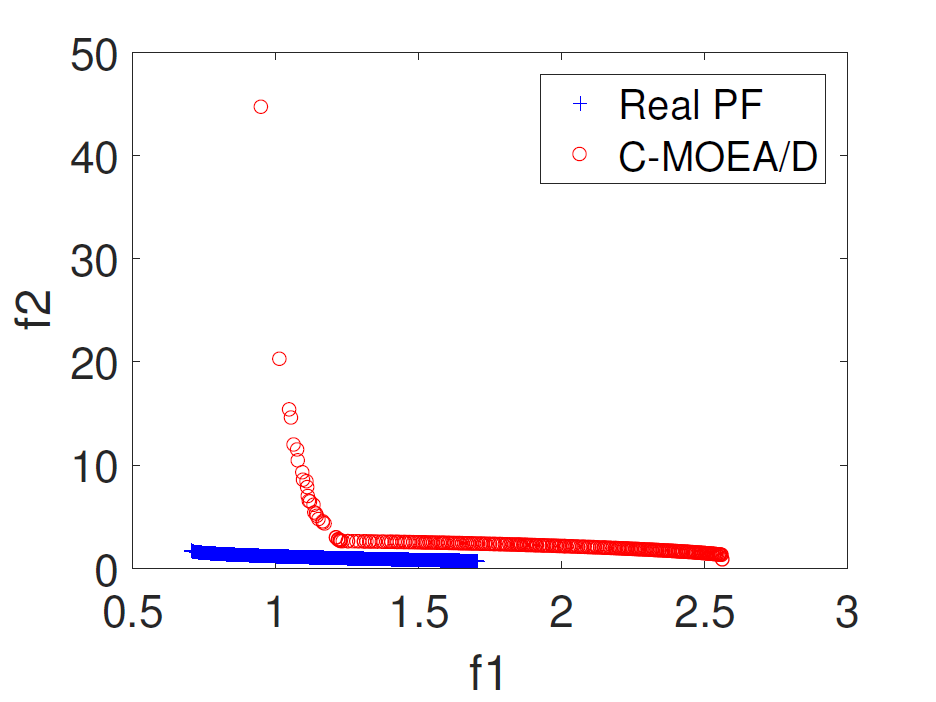}\\
\end{minipage}
\begin{minipage}[t]{0.25\linewidth}
\includegraphics[width = 3.3cm]{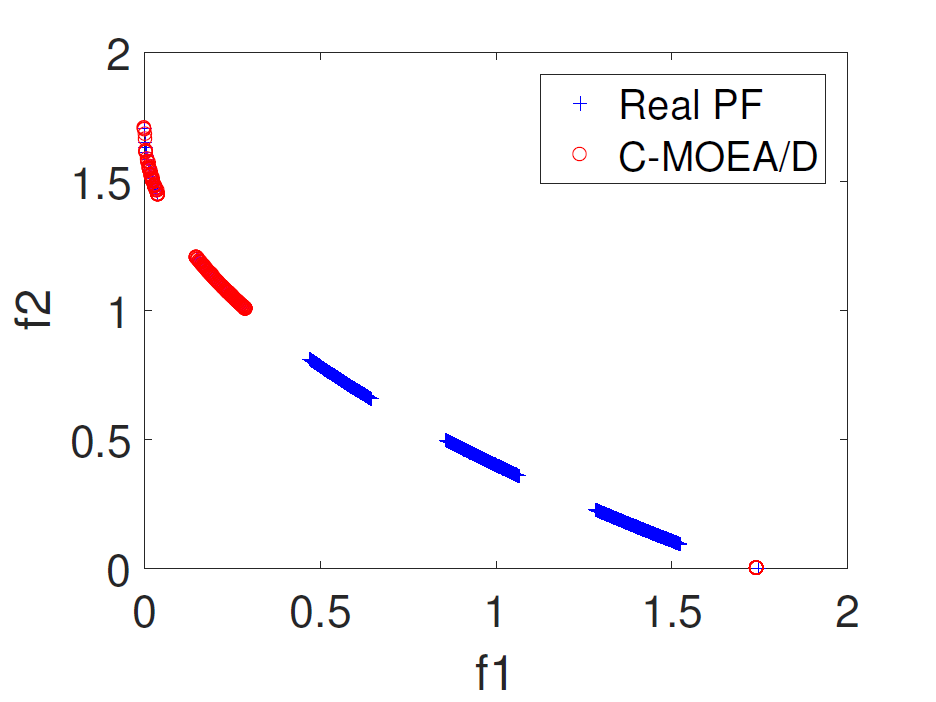}\\
\end{minipage}
\begin{minipage}[t]{0.25\linewidth}
\includegraphics[width = 3.3cm]{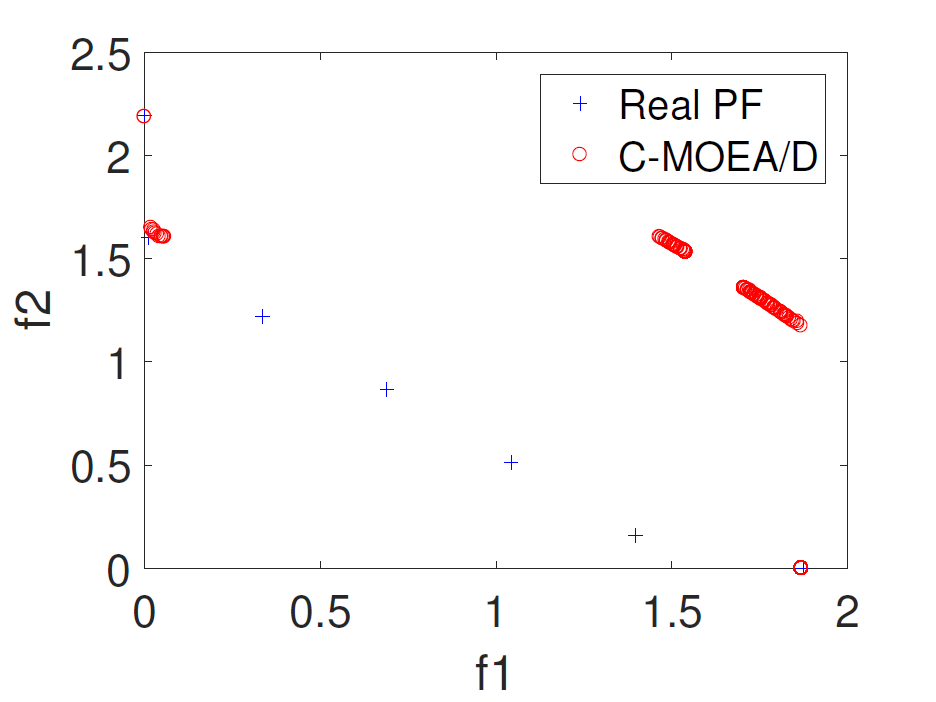}\\
\end{minipage}
\end{tabular}

\begin{tabular}{cc}
\begin{minipage}[t]{0.25\linewidth}
\includegraphics[width = 3.3cm]{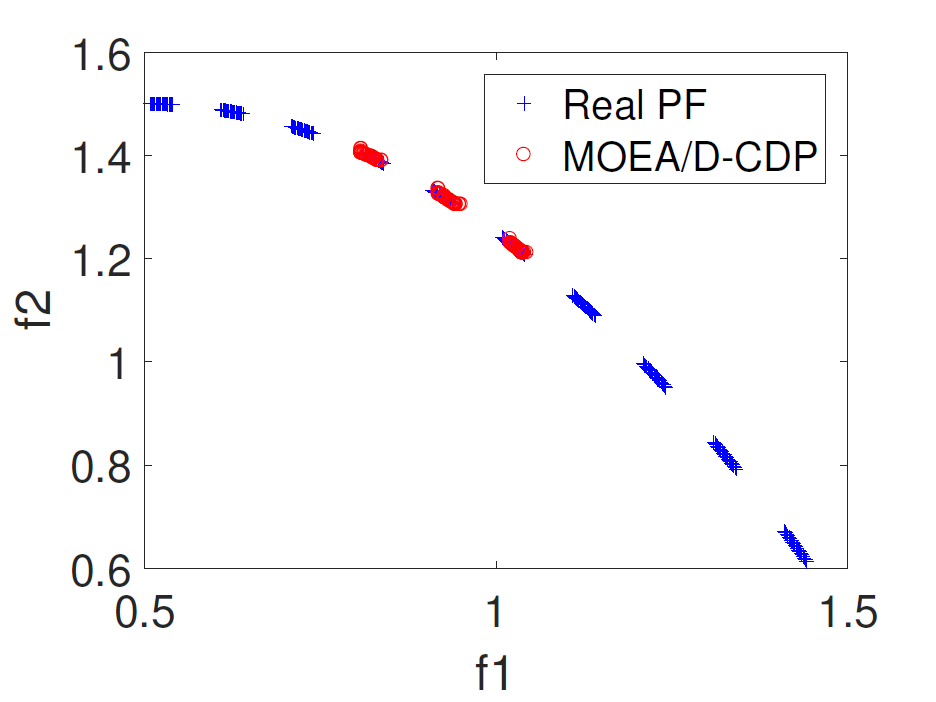}\\
\end{minipage}
\begin{minipage}[t]{0.25\linewidth}
\includegraphics[width = 3.3cm]{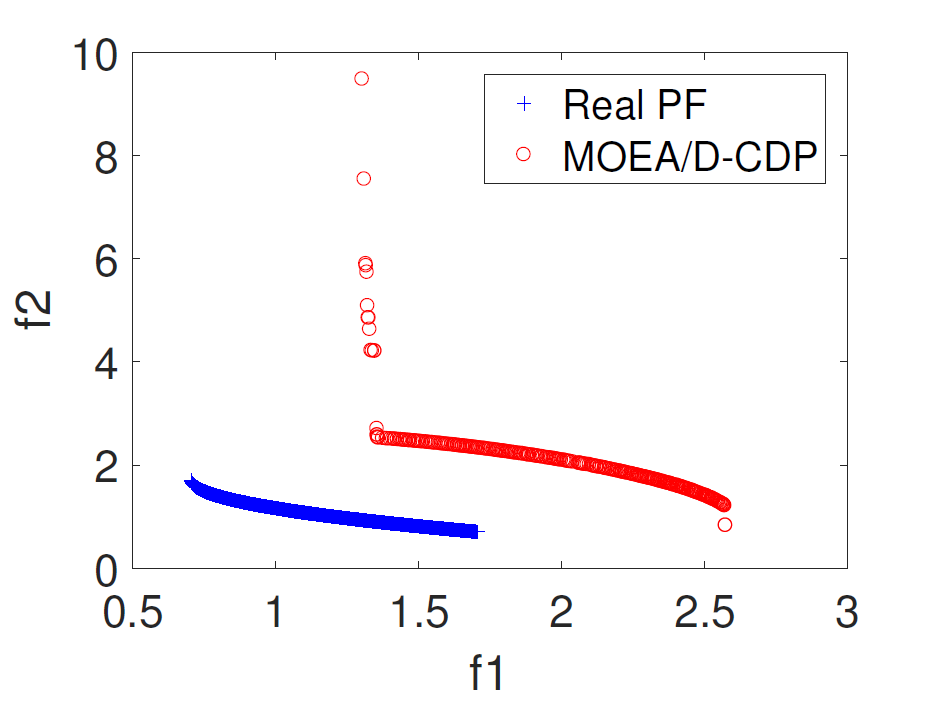}\\
\end{minipage}
\begin{minipage}[t]{0.25\linewidth}
\includegraphics[width = 3.3cm]{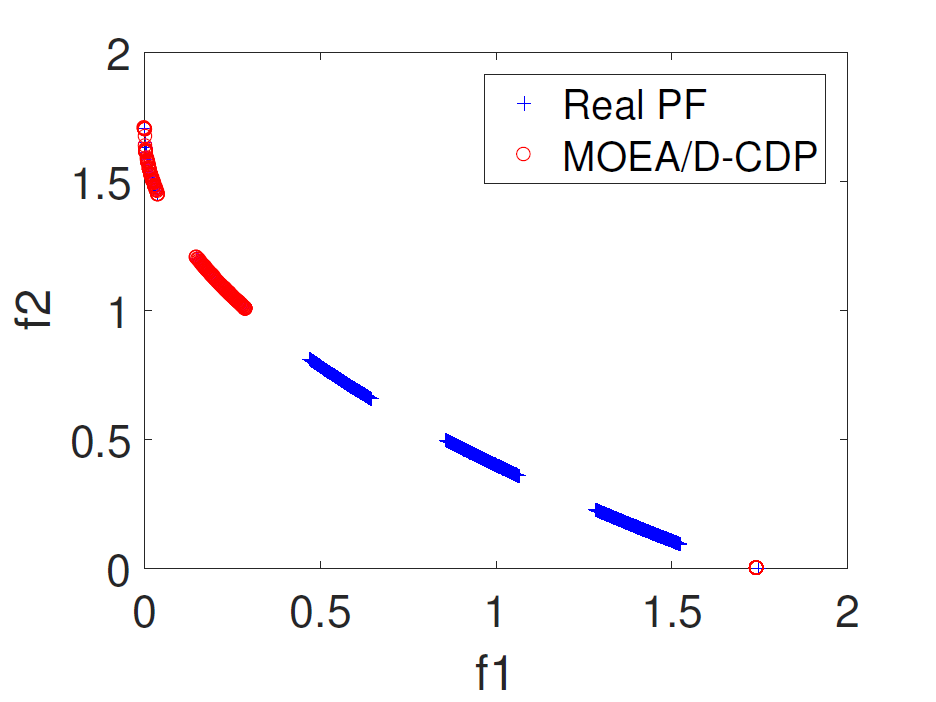}\\
\end{minipage}
\begin{minipage}[t]{0.25\linewidth}
\includegraphics[width = 3.3cm]{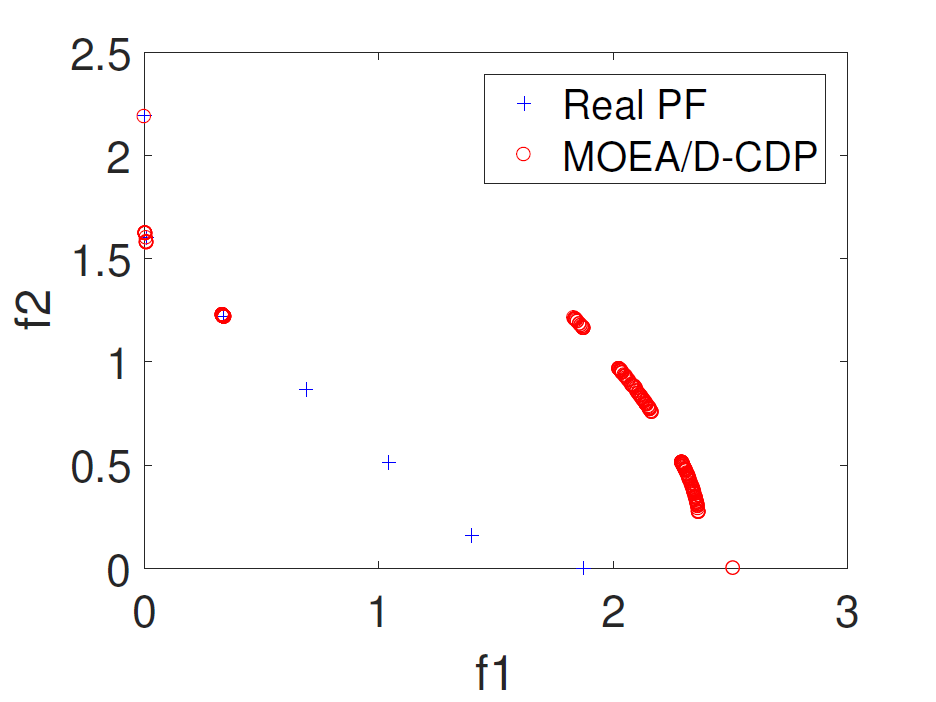}\\
\end{minipage}
\end{tabular}

\begin{tabular}{cc}
\begin{minipage}[t]{0.25\linewidth}
\includegraphics[width = 3.3cm]{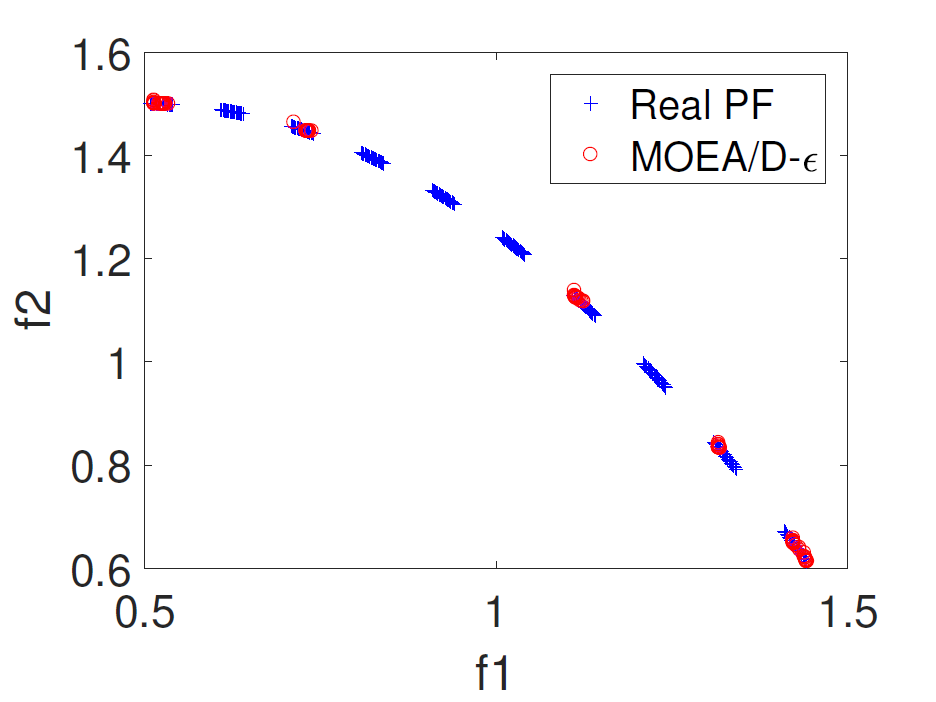}\\
\end{minipage}
\begin{minipage}[t]{0.25\linewidth}
\includegraphics[width = 3.3cm]{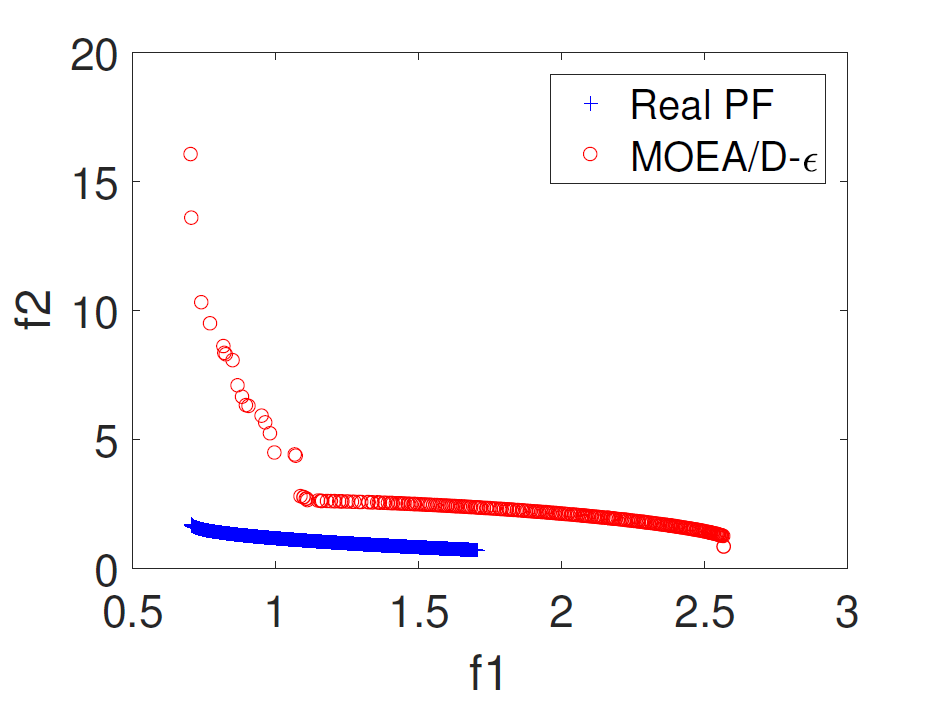}\\
\end{minipage}
\begin{minipage}[t]{0.25\linewidth}
\includegraphics[width = 3.3cm]{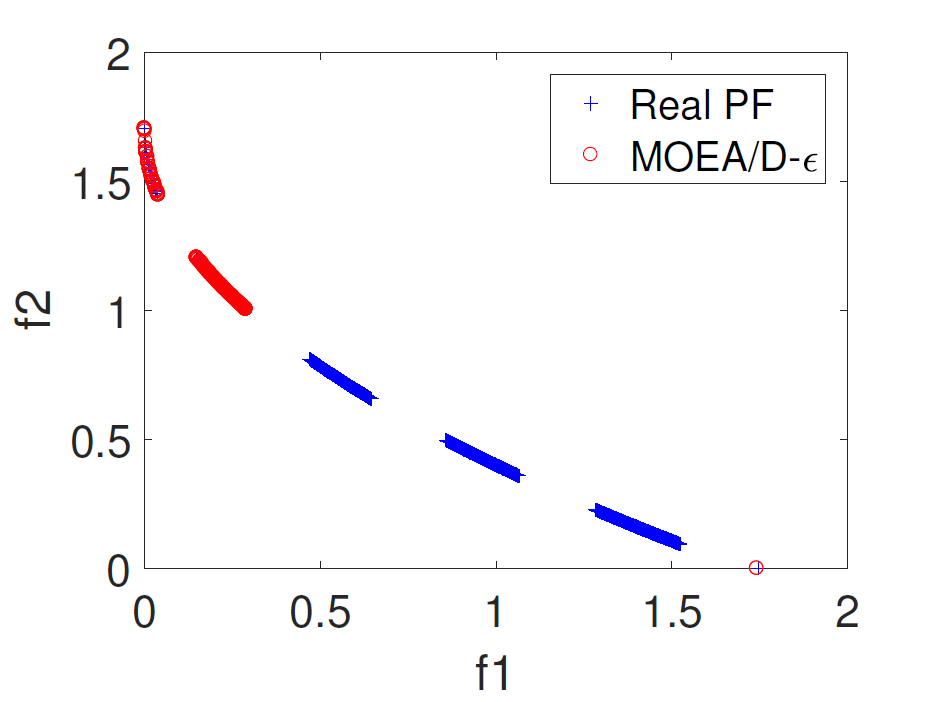}\\
\end{minipage}
\begin{minipage}[t]{0.25\linewidth}
\includegraphics[width = 3.3cm]{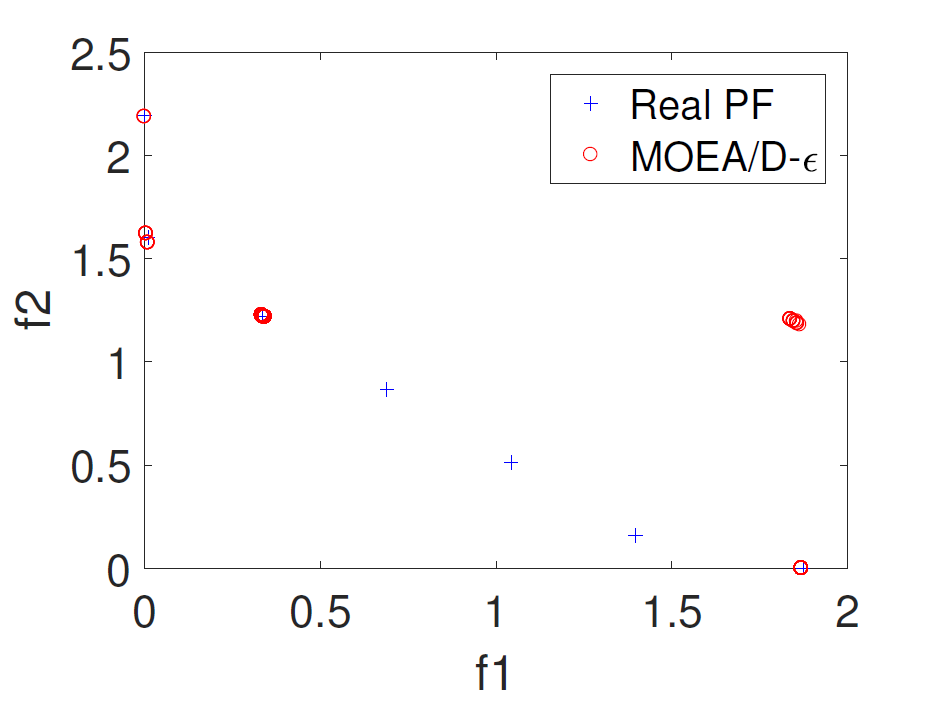}\\
\end{minipage}
\end{tabular}

\begin{tabular}{cc}
\begin{minipage}[t]{0.25\linewidth}
\includegraphics[width = 3.3cm]{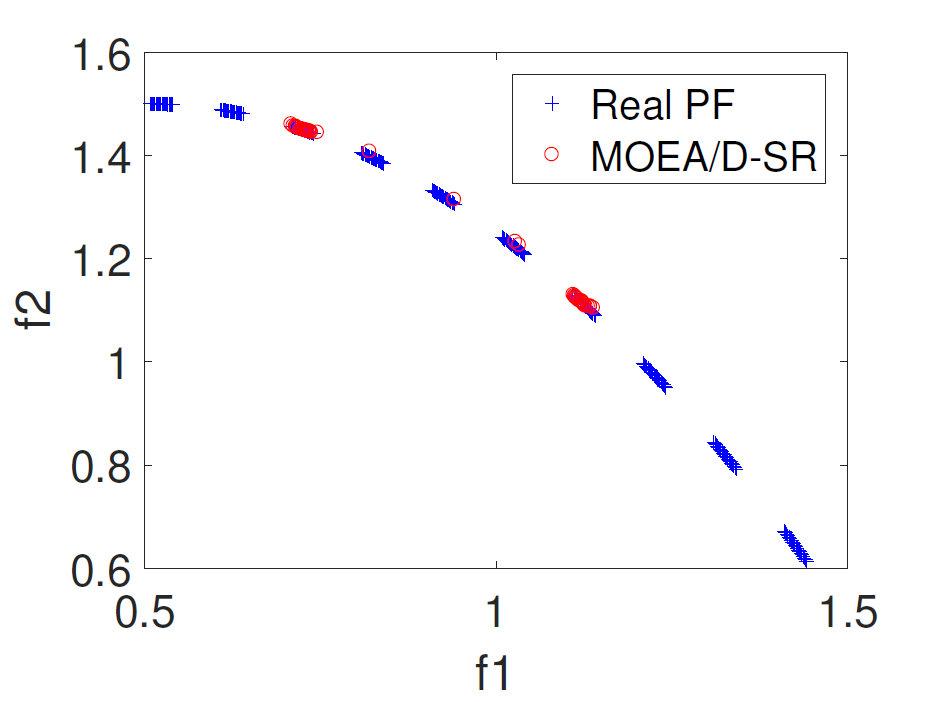}\\
\centering{\scriptsize{(a) LIR-CMOP3}}
\end{minipage}
\begin{minipage}[t]{0.25\linewidth}
\includegraphics[width = 3.3cm]{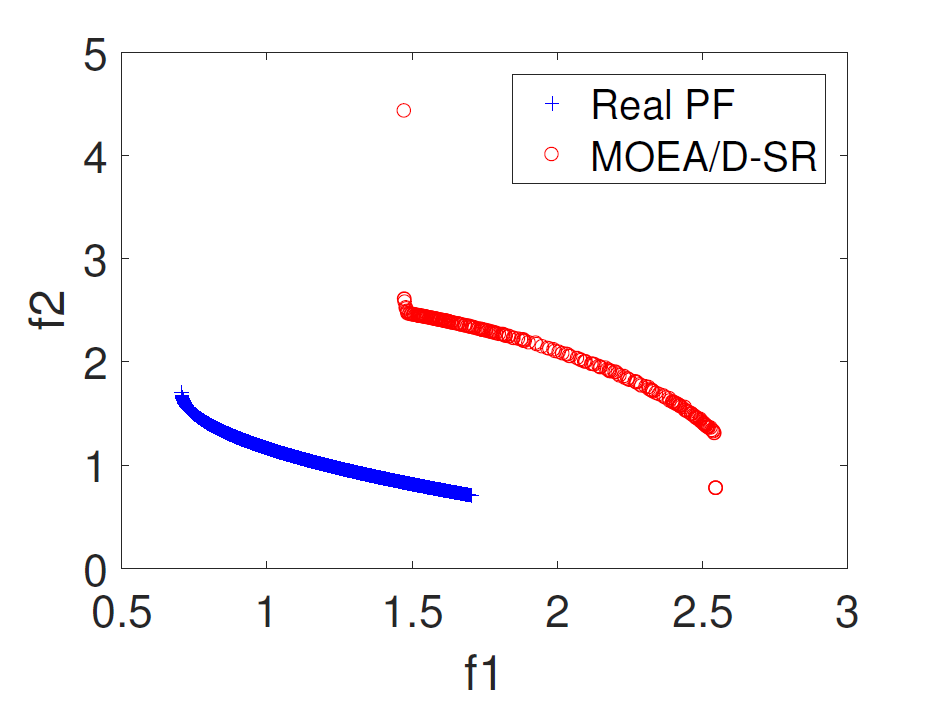}\\
\centering{\scriptsize{(b) LIR-CMOP5}}
\end{minipage}
\begin{minipage}[t]{0.25\linewidth}
\includegraphics[width = 3.3cm]{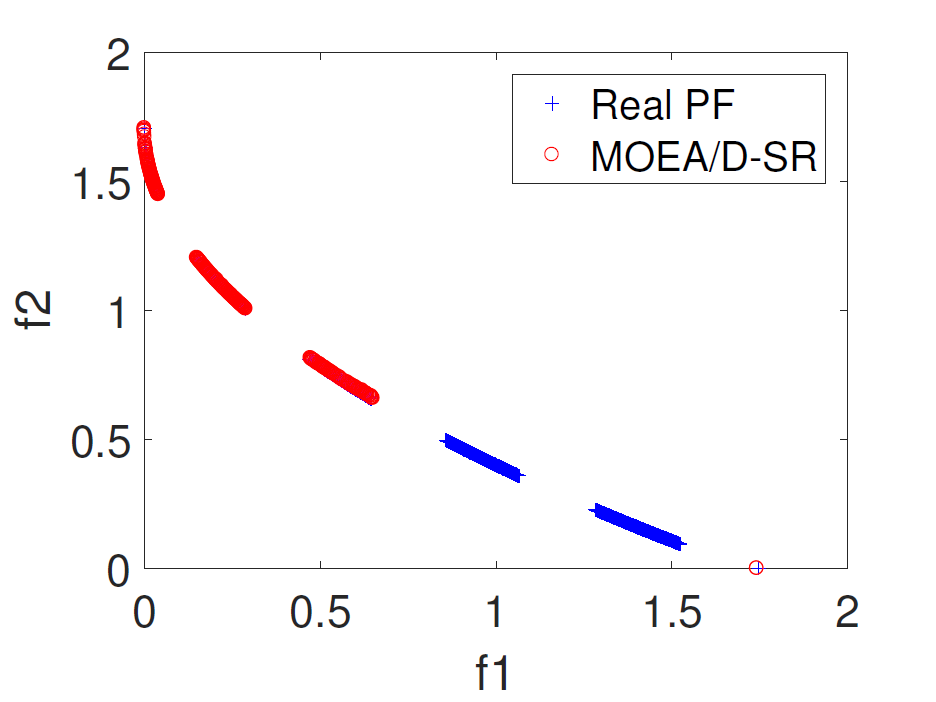}\\
\centering{\scriptsize{(c) LIR-CMOP10}}
\end{minipage}
\begin{minipage}[t]{0.25\linewidth}
\includegraphics[width = 3.3cm]{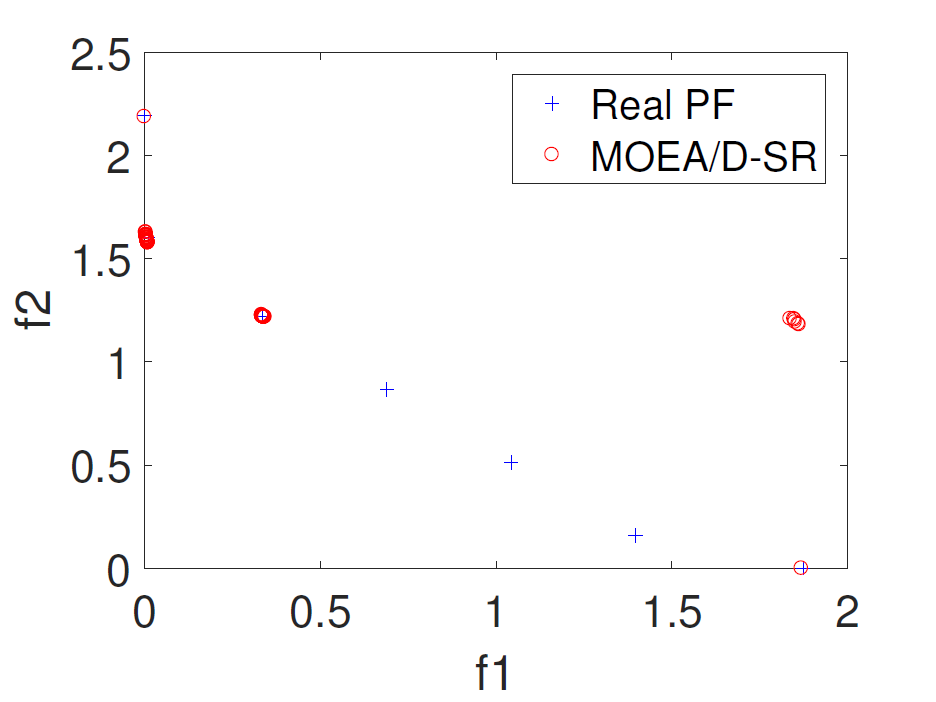}\\
\centering{\scriptsize{(d) LIR-CMOP11}}
\end{minipage}
\end{tabular}

\centering{\caption{The non-dominated solutions achieved by each algorithm with the median $IGD$ in the 30 independent runs for LIR-CMOP3, LIR-CMOP5, LIR-CMOP10 and LIR-CMOP11.}\label{fig:lir-cmop-selected}}
\end{figure*}

Fig. \ref{fig:lir-cmop-selected} (a) shows the final external archives achieved by MOEA/D-ACDP and other four CMOEAs with the median $IGD$ values on LIR-CMOP3 during 30 independent runs. It is obvious that MOEA/D-ACDP can almost converge to the whole real PF and has the best diversity among the five CMOEAs.

In Fig. \ref{fig:lir-cmop-selected} (b), the results of each CMOEA with the median $IGD$ values on LIR-CMOP5 during 30 independent runs are shown. The external archive achieved by MOEA/D-ACDP covers the real PF. However, the other four CMOEAs are trapped into local optimum. As shown in Fig. \ref{fig:lir-cmop-selected} (c), for LIR-CMOP10, MOEA/D-ACDP performs the best in terms of convergency. In Fig. \ref{fig:lir-cmop-selected} (d), for LIR-CMOP11, it shows that MOEA/D-ACDP can get the most of the PF, but the other four algorithms can only achieve a few parts of the PF.

It is worthwhile to point out that for three-objective test instances (LIR-CMOP13 and LIR-CMOP14), MOEA/D-ACDP also performs significantly better than the other four CMOEAs.

Based on the above performance comparison on the fourteen test instances LIR-CMOP1-14, it is clear that MOEA/D-ACDP outperforms the other four decomposition-based CMOEAs on most of cases.

A common feature of the above test instances LIR-CMOPs is that they all have large infeasible regions in their objective space. The experimental results demonstrate that the proposed ACDP method can deal with CMOPs well by taking advantage of angle information of the working population.

\begin{table*}[htbp]
   \centering
  \caption{$IGD$ results of MOEA/D-ACDP and the other four CMOEAs on LIR-CMOP1-14 test instances}
    \resizebox{!}{4.2cm}{
    \begin{tabular}{c|c|ccccc}
    \toprule
    \multicolumn{2}{c|}{Test Instances} & MOEA/D-ACDP & C-MOEA/D & MOEA/D-CDP & MOEA/D-Epsilon & MOEA/D-SR \\
    \hline
    \multirow{2}[0]{*}{LIR-CMOP1} & mean  & 5.159E-02 & 1.591E-01$^{\dag}$ & 1.348E-01$^{\dag}$ & 8.234E-02$^{\dag}$ & \textbf{4.406E-02} \\
          & std   & 1.815E-02 & 3.534E-02 & 5.996E-02 & 5.321E-02 & 3.360E-02 \\
    \hline
    \multirow{2}[0]{*}{LIR-CMOP2} & mean  & 2.269E-02 & 1.462E-01$^{\dag}$ & 1.549E-01$^{\dag}$ & 4.708E-02$^{\dag}$& \textbf{2.057E-02} \\
          & std   & 9.418E-03 & 4.141E-02 & 2.966E-02 & 1.339E-02 & 1.072E-02 \\
    \hline
    \multirow{2}[0]{*}{LIR-CMOP3} & mean  & \textbf{4.659E-02} & 2.309E-01$^{\dag}$ & 2.268E-01$^{\dag}$ & 7.858E-02$^{\dag}$ & 1.529E-01$^{\dag}$ \\
          & std   & 1.850E-02 & 4.135E-02 & 4.403E-02 & 2.978E-02 & 7.688E-02 \\
    \hline
    \multirow{2}[0]{*}{LIR-CMOP4} & mean  & \textbf{2.784E-02} &2.080E-01$^{\dag}$ &2.188E-01$^{\dag}$ &5.662E-02$^{\dag}$ &2.038E-01$^{\dag}$ \\
          & std   & 1.477E-02 &4.197E-02 &3.766E-02 &3.366E-02 &7.907E-02 \\
    \hline
    \multirow{2}[0]{*}{LIR-CMOP5} & mean  & \textbf{1.771E-02} &1.162E+00$^{\dag}$ &1.207E+00$^{\dag}$ &1.201E+00$^{\dag}$ &1.123E+00$^{\dag}$ \\
          & std   & 2.965E-02 &2.180E-01 &1.660E-02 &1.963E-02 &2.842E-01 \\
    \hline
    \multirow{2}[0]{*}{LIR-CMOP6} & mean  & \textbf{1.757E-01} &1.265E+00$^{\dag}$ &1.303E+00$^{\dag}$ &1.231E+00$^{\dag}$ &1.175E+00$^{\dag}$ \\
          & std   & 4.129E-02 &3.067E-01 &2.319E-01 &3.602E-01 &3.967E-01 \\
    \hline
    \multirow{2}[0]{*}{LIR-CMOP7} & mean  & \textbf{1.408E-01} &1.620E+00$^{\dag}$ &1.623E+00$^{\dag}$ &1.568E+00$^{\dag}$ &1.136E+00$^{\dag}$ \\
          & std   & 4.385E-02 &3.036E-01 &2.905E-01 &4.101E-01 &7.315E-01 \\
    \hline
    \multirow{2}[0]{*}{LIR-CMOP8} & mean  & \textbf{1.812E-01} &1.607E+00$^{\dag}$ &1.631E+00$^{\dag}$ &1.577E+00$^{\dag}$ &1.369E+00$^{\dag}$ \\
          & std   & 4.854E-02 &2.680E-01 &2.464E-01 &3.767E-01 &5.735E-01 \\
    \hline
    \multirow{2}[0]{*}{LIR-CMOP9} & mean  & \textbf{3.595E-01} &4.981E-01$^{\dag}$ &4.868E-01$^{\dag}$ &4.962E-01$^{\dag}$ &4.813E-01$^{\dag}$ \\
          & std   & 5.345E-02 &6.991E-02 &5.372E-02 &6.987E-02 &4.571E-02 \\
    \hline
    \multirow{2}[0]{*}{LIR-CMOP10} & mean  & \textbf{1.388E-01} &3.775E-01$^{\dag}$ &3.774E-01$^{\dag}$ &3.257E-01$^{\dag}$ &2.821E-01$^{\dag}$ \\
          & std   & 1.148E-01 &7.446E-02 &6.858E-02 &9.833E-02 &1.135E-01 \\
    \hline
    \multirow{2}[0]{*}{LIR-CMOP11} & mean  & \textbf{1.318E-01} &4.422E-01$^{\dag}$ &4.662E-01$^{\dag}$ &4.154E-01$^{\dag}$ &3.489E-01$^{\dag}$ \\
          & std   & 4.487E-02 &1.759E-01 &1.439E-01 &1.508E-01 &1.129E-01 \\
    \hline
    \multirow{2}[0]{*}{LIR-CMOP12} & mean  & \textbf{1.497E-01} &3.597E-01$^{\dag}$ &3.236E-01$^{\dag}$ &3.680E-01$^{\dag}$ &3.012E-01$^{\dag}$ \\
          & std   & 9.985E-03 &1.074E-01 &1.023E-01 &8.664E-02 &8.989E-02 \\
    \hline
    \multirow{2}[0]{*}{LIR-CMOP13} & mean  & \textbf{7.414E-02} &1.266E+00$^{\dag}$ &1.289E+00$^{\dag}$ &1.183E+00$^{\dag}$ &1.093E+00$^{\dag}$ \\
          & std   & 2.727E-03 &2.173E-01 &6.321E-02 &3.456E-01 &4.269E-01 \\
    \hline
    \multirow{2}[0]{*}{LIR-CMOP14} & mean  & \textbf{6.732E-02} &1.235E+00$^{\dag}$ &1.103E+00$^{\dag}$ &1.127E+00$^{\dag}$ &1.143E+00$^{\dag}$ \\
          & std   & 1.918E-03 &1.209E-01 &3.857E-01 &3.329E-01 &3.002E-01 \\
    \bottomrule
    \end{tabular}}%
  \label{tab:lir-cmop-igd} \\
  \footnotesize Wilcoxon’s rank sum test at a 0.05 significance level is performed between MOEA/D-ACDP and each of the other four CMOEAs. $\dag$ and $\ddag$ denote that the performance of the corresponding algorithm is significantly worse than or better than that of MOEA/D-ACDP, respectively. The best mean is highlighted in boldface.
\end{table*}%

\begin{table*}[htbp]
  \centering
  \caption{$HV$ results of MOEA/D-ACDP and the other four CMOEAs on LIR-CMOP1-14 test instances}
    \resizebox{!}{4.2cm}{
    \begin{tabular}{c|c|ccccc}
    \toprule
    \multicolumn{2}{c|}{Test Instances} & MOEA/D-ACDP & C-MOEA/D & MOEA/D-CDP & MOEA/D-Epsilon & MOEA/D-SR \\
    \hline
    \multirow{2}[0]{*}{LIR-CMOP1} & mean     &1.365E+00 &9.499E-01$^{\dag}$ &1.009E+00$^{\dag}$ &1.353E+00$^{\dag}$ &\textbf{1.376E+00} \\
          & std     & 2.493E-02 &7.038E-02 &1.298E-01 &4.417E-02 &3.974E-02 \\
    \hline
    \multirow{2}[0]{*}{LIR-CMOP2} & mean     & \textbf{1.737E+01} &1.395E+01$^{\dag}$ &1.374E+01$^{\dag}$ &1.705E+01$^{\dag}$ &1.736E+01$^{\dag}$ \\
          & std     & 1.306E-02 &8.154E-02 &6.160E-02 &1.693E-02 &1.890E-02 \\
    \hline
    \multirow{2}[0]{*}{LIR-CMOP3} & mean     & \textbf{1.188E+00} &7.558E-01$^{\dag}$ &7.600E-01$^{\dag}$ &1.184E+00$^{\dag}$ &9.313E-01$^{\dag}$ \\
          & std    & 4.929E-02 &5.730E-02 &5.809E-02 &2.898E-02 &1.620E-01 \\
    \hline
    \multirow{2}[0]{*}{LIR-CMOP4} & mean     & \textbf{1.421E+00} &1.069E+00$^{\dag}$ &1.051E+00$^{\dag}$ &1.390E+00$^{\dag}$ &1.089E+00$^{\dag}$ \\
          & std     & 1.946E-02 &6.952E-02 &5.462E-02 &4.405E-02 &1.360E-01 \\
    \hline
    \multirow{2}[0]{*}{LIR-CMOP5} & mean     & \textbf{1.903E+00} &1.192E-01$^{\dag}$ &5.805E-02$^{\dag}$ &5.829E-02$^{\dag}$ &1.707E-01$^{\dag}$ \\
          & std    & 5.658E-02 &3.352E-01 &4.042E-04 &2.022E-04 &4.442E-01 \\
    \hline
    \multirow{2}[0]{*}{LIR-CMOP6} & mean    & \textbf{1.280E+00} &7.863E-02$^{\dag}$ &4.312E-02$^{\dag}$ &1.325E-01$^{\dag}$ &1.682E-01$^{\dag}$ \\
          & std    & 4.613E-02 &3.011E-01 &2.362E-01 &4.251E-01 &4.061E-01 \\
    \hline
    \multirow{2}[0]{*}{LIR-CMOP7} & mean    & \textbf{3.408E+00} &2.990E-01$^{\dag}$ &2.886E-01$^{\dag}$ &4.055E-01$^{\dag}$ &1.313E+00$^{\dag}$\\
          & std    & 1.409E-01 &6.927E-01 &6.348E-01 &8.879E-01 &1.567E+00 \\
    \hline
    \multirow{2}[0]{*}{LIR-CMOP8} & mean    & \textbf{3.330E+00} &3.246E-01$^{\dag}$ &2.695E-01$^{\dag}$ &3.859E-01$^{\dag}$ &8.287E-01$^{\dag}$ \\
          & std    & 1.461E-01 &5.878E-01 &5.297E-01 &8.166E-01 &1.244E+00 \\
    \hline
    \multirow{2}[0]{*}{LIR-CMOP9} & mean    & \textbf{4.080E+00} &3.715E+00$^{\dag}$ &3.755E+00$^{\dag}$ &3.724E+00$^{\dag}$ &3.752E+00$^{\dag}$ \\
          & std    & 9.501E-02 &2.079E-01 &1.600E-01 &2.033E-01 &1.142E-01 \\
    \hline
    \multirow{2}[0]{*}{LIR-CMOP10} & mean    & \textbf{3.755E+00} &3.274E+00$^{\dag}$ &3.268E+00$^{\dag}$ &3.385E+00$^{\dag}$ &3.477E+00$^{\dag}$ \\
          & std    & 2.208E-01 &1.623E-01 &1.416E-01 &2.122E-01 &2.383E-01 \\
    \hline
    \multirow{2}[0]{*}{LIR-CMOP11} & mean    & \textbf{5.004E+00} &3.937E+00$^{\dag}$ &3.842E+00$^{\dag}$ &4.038E+00$^{\dag}$ &4.274E+00$^{\dag}$ \\
          & std    & 1.564E-01 &6.479E-01 &5.507E-01 &5.727E-01 &4.463E-01 \\
    \hline
    \multirow{2}[0]{*}{LIR-CMOP12} & mean    & \textbf{6.713E+00} &5.977E+00$^{\dag}$ &6.134E+00$^{\dag}$ &6.010E+00$^{\dag}$ &6.240E+00$^{\dag}$ \\
          & std    & 05.874E-02 &3.855E-01 &3.617E-01 &3.074E-01 &2.950E-01 \\
    \hline
    \multirow{2}[0]{*}{LIR-CMOP13} & mean    & \textbf{7.897E+00} &6.444E-01$^{\dag}$ &4.728E-01$^{\dag}$ &1.092E+00$^{\dag}$ &1.513E+00$^{\dag}$ \\
          & std    & 2.943E-02 &1.317E+00 &2.689E-01 &2.0522E+00 &2.422E+00 \\
    \hline
    \multirow{2}[0]{*}{LIR-CMOP14} & mean    & \textbf{8.641E+00} &7.766E-01$^{\dag}$ &1.627E+00$^{\dag}$ &1.430E+00$^{\dag}$ &1.269E+00$^{\dag}$ \\
          & std    & 1.546E-02 &6.140E-01 &2.473E+00 &2.095E+00 &1.919E+00 \\
    \bottomrule
    \end{tabular}}%
    \label{tab:lir-cmop-hv}\\
    \footnotesize Wilcoxon’s rank sum test at a 0.05 significance level is performed between MOEA/D-ACDP and each of the other four CMOEAs. $\dag$ and $\ddag$ denotes that the performance of the corresponding algorithm is significantly worse than or better than that of MOEA/D-ACDP, respectively. The best mean is highlighted in boldface.
\end{table*}%

\subsubsection{Performance Evaluation on I-beam Optimization Problem}
\label{sec:4.5.2}
The experimental results of $HV$ values of MOEA/D-ACDP and the four other CMOEAs on the I-beam optimization problem are shown in Table \ref{tab:Ibeam-HV}. It can be observed that MOEA/D-ACDP significantly outperforms the compared CMOEAs on this engineering problem.

\begin{figure*}[htbp]
\begin{tabular}{cc}
\begin{minipage}[t]{0.33\linewidth}
\includegraphics[width = 4cm]{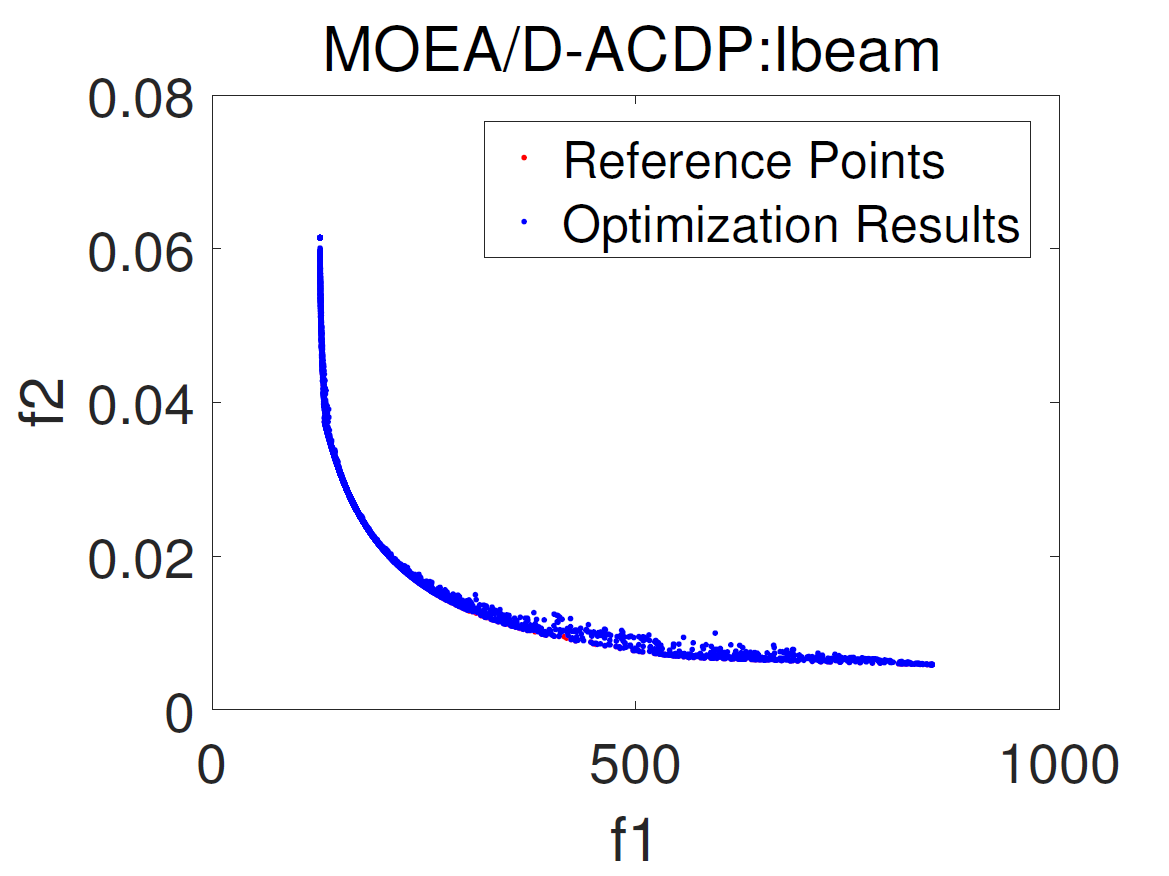}\\
\centering{\scriptsize{(a) MOEA/D-ACDP}}
\end{minipage}
\begin{minipage}[t]{0.33\linewidth}
\includegraphics[width = 4cm]{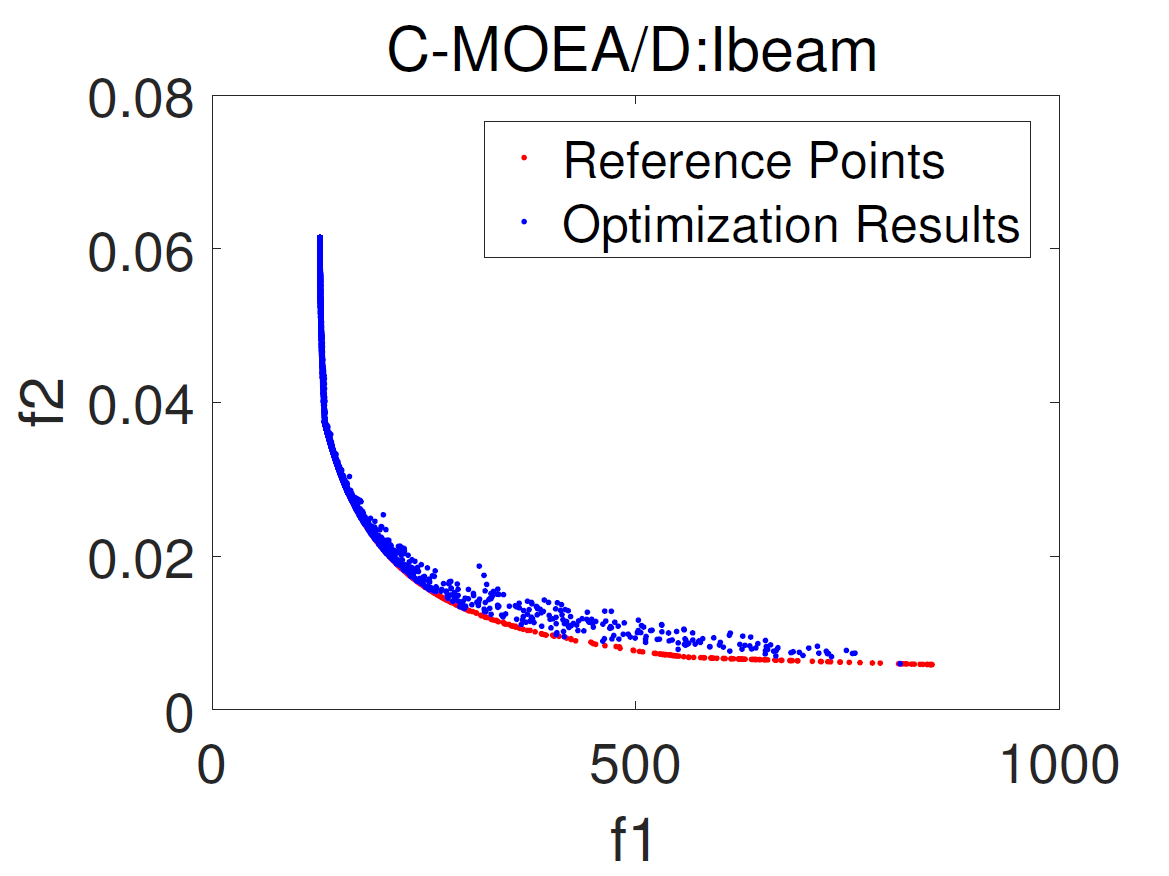}\\
\centering{\scriptsize{(b) C-MOEA/D}}
\end{minipage}
\begin{minipage}[t]{0.33\linewidth}
\includegraphics[width = 4cm]{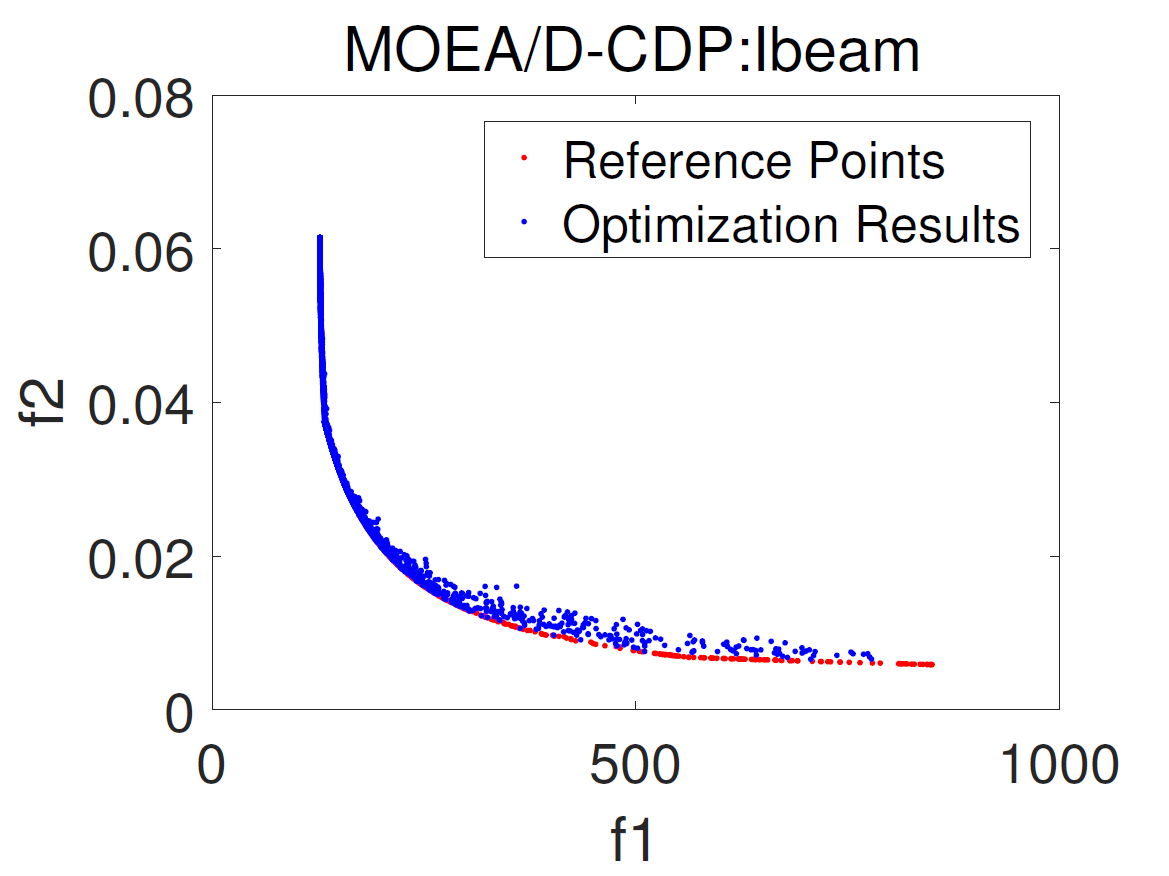}\\
\centering{\scriptsize{(c) MOEA/D-CDP}}
\end{minipage}
\end{tabular}

\begin{tabular}{cc}
\begin{minipage}[t]{0.33\linewidth}
\includegraphics[width = 4cm]{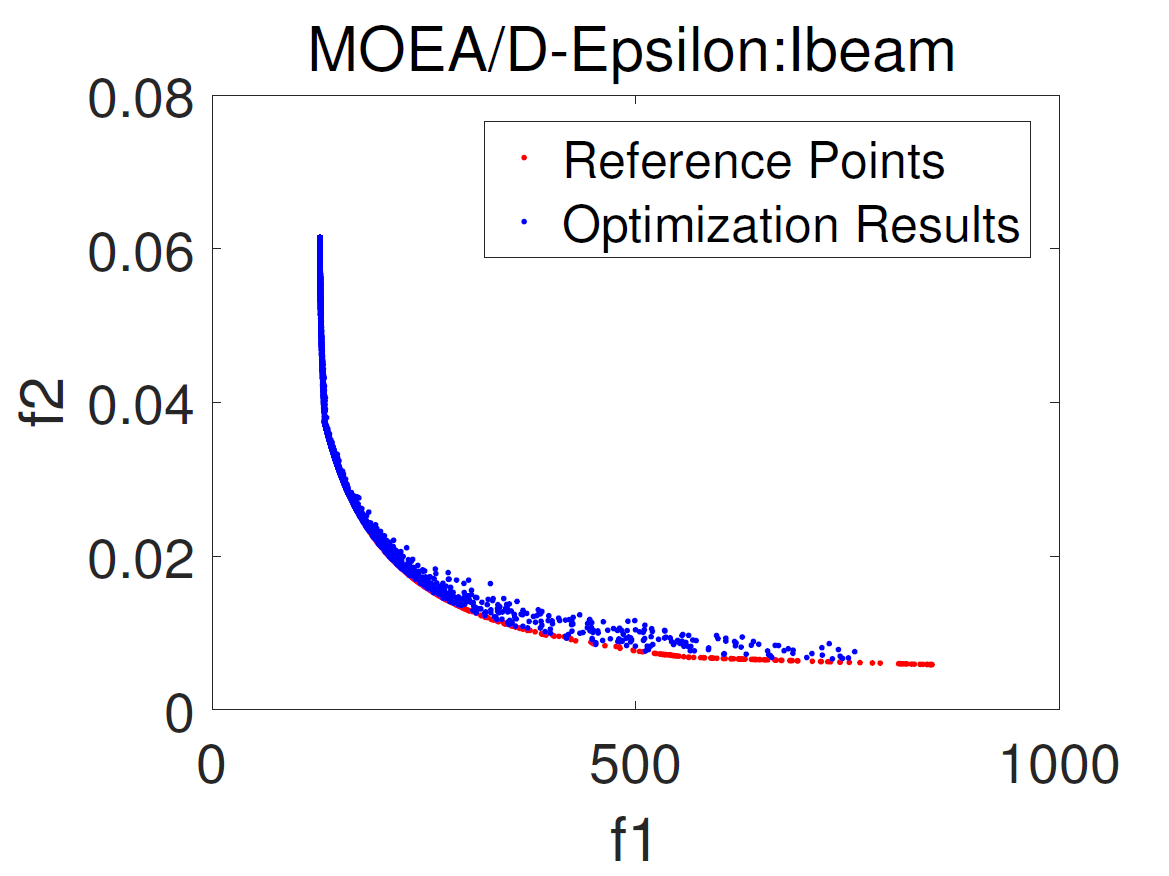}\\
\centering{\scriptsize{(d) MOEA/D-Epsilon}}
\end{minipage}
\begin{minipage}[t]{0.33\linewidth}
\includegraphics[width = 4cm]{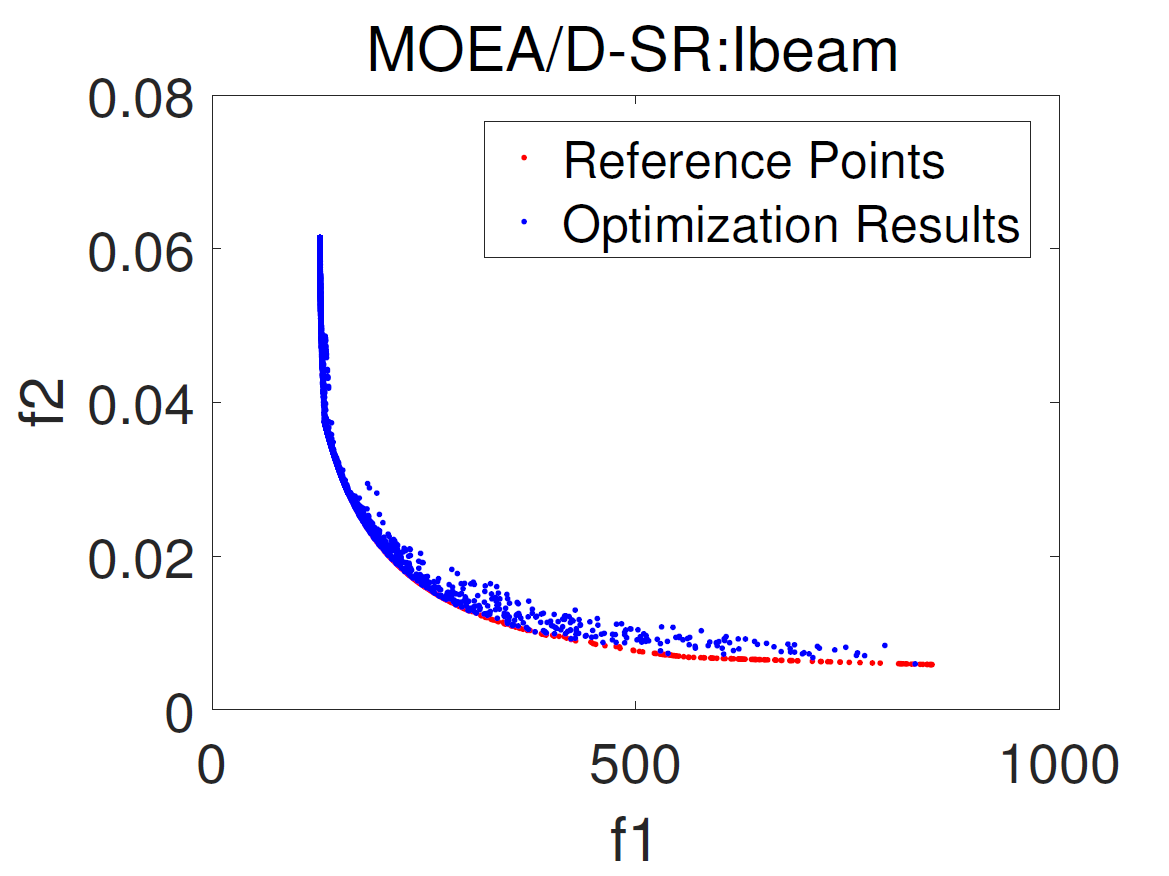}\\
\centering{\scriptsize{(e) MOEA/D-SR}}
\end{minipage}

\begin{minipage}[t]{0.33\linewidth}
\includegraphics[width = 4cm,height = 3cm]{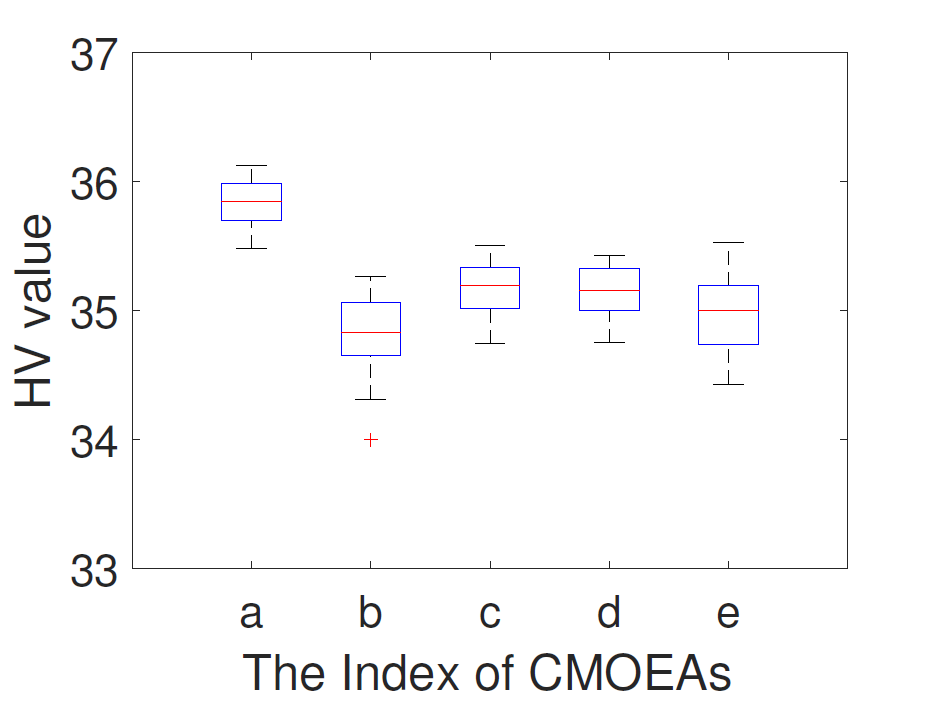}\\
\centering{\scriptsize{(f) The box plots of each CMOEA}}
\end{minipage}

\end{tabular}

\caption{The non-dominated solutions achieved by each algorithm during the 30 independent runs are plotted in (a)-(e). In (f), the box plots of each CMOEA are plotted.}
\label{Fig:Ibeam_result}
\end{figure*}

To further study the superiority of the proposed method MOEA/D-ACDP, the non-dominated solutions achieved by each CMOEA during the 30 independent runs are plotted in Fig. \ref{Fig:Ibeam_result} (a)-(e).The non-dominated set of all the above solutions generates a set of ideal reference points. It is clear that the external archive obtained by MOEA/D-ACDP has a better performance of convergence. The box plot of $HV$ values of the five CMOEAs is shown in Fig. \ref{Fig:Ibeam_result} (f), which further verifies that MOEA/D-ACDP outperforms the other four CMOEAs on the I-beam optimization problem.

\label{sec:6.2.2}

\begin{table*}[htbp]
  \centering
  \caption{$HV$ results of MOEA/D-ACDP and the other four CMOEAs on the I-Beam optimization problem}
    \resizebox{!}{0.6cm}{
    \begin{tabular}{c|c|ccccc}
    \toprule
    {Test Instances} & MOEA/D-ACDP & C-MOEA/D & MOEA/D-CDP & MOEA/D-Epsilon & MOEA/D-SR \\
    \hline
    mean  & \textbf{3.583E+01} & 3.481E+01$^{\dag}$ & 3.518E+01$^{\dag}$
    & 3.514E+03$^{\dag}$ & 3.477E+03$^{\dag}$ \\
    std   & 1.950E-01 & 2.968E-01 & 2.078E-01 & 2.034E-01 & 1.248E+00 \\
    \bottomrule
    \end{tabular}}%
  \label{tab:Ibeam-HV} \\
  \footnotesize Wilcoxon’s rank sum test at a 0.05 significance level is performed between MOEA/D-ACDP and each of the other four CMOEAs. $\dag$ and $\ddag$ denote that the performance of the corresponding algorithm is significantly worse than or better than that of MOEA/D-ACDP, respectively. The best mean is highlighted in boldface.
\end{table*}%

\section{Conclusions}
\label{sec:5}
This paper proposes a new constraint-handling mechanism named ACDP. It utilizes the angle information of any two solutions to dynamically maintain the diversity of the population during the evolutionary process.
A set of CMOP instances named LIR-CMOP1-14 are tested. All the test instances have large infeasible regions in their objective space, which make general CMOEAs difficult to achieve the real PFs. Compared with the other four popular CMOEAs, the proposed algorithm can help the population to go across large infeasible regions more effectively. Additionally, the experimental results demonstrate that the proposed algorithm can work well in the real-world engineering problem. Thus, we can conclude that MOEA/D-ACDP outperforms the other four CMOEAs when CMOPs. In summary, MOEA/D-ACDP has following advantages:

\begin{itemize}
  \item The proposed MOEA/D-ACDP utilizes the angle information of solutions to maintain the diversity of the population for CMOPs.
  \item MOEA/D-ACDP enhances the convergence to PF by exploring feasible and infeasible regions simultaneously during the evolutionary process, instead of wasting the useful information of the infeasible solutions.

\end{itemize}

Future work will focus on novel constraint-handling mechanisms to solve CMOPs. A study on developing new mechanisms of mining more useful information during the evolutionary process to further improve the performance of the proposed algorithm will be conducted.
\section*{Acknowledgment}
This work was supported in part by the National Natural Science Foundation of China under Grant (61175073,
61300159, 61332002, 51375287) , the Guangdong Key Laboratory of Digital signal and Image Processing, the Science
and Technology Planning Project of Guangdong Province (2013B011304002) and the Project of Educational Commission of Guangdong Province, China 2015KGJHZ014).

\section*{References}

\bibliography{acdp}

\end{document}